\documentclass[10pt,twocolumn,letterpaper]{article}

\usepackage[pagenumbers]{cvpr} 

\usepackage{graphicx}
\usepackage{amsmath}
\usepackage{amssymb}
\usepackage{booktabs}
\usepackage{comment}

\graphicspath{ {images/} }
\usepackage{soul}
\usepackage{xcolor}
\usepackage{xspace}

\usepackage{pifont}

\newcommand{\mypar}[1]{\vspace{1mm}\noindent{\bf #1}}

\definecolor{asparagus}{rgb}{0.53, 0.66, 0.42}
\definecolor{armygreen}{rgb}{0.29, 0.33, 0.13}
\definecolor{awesome}{rgb}{1.0, 0.13, 0.32}
\definecolor{applegreen}{rgb}{0.55, 0.71, 0.0}

\newcommand{\unet}{Parallel-UNet\xspace}

\usepackage[pagebackref,breaklinks,colorlinks]{hyperref}

\usepackage[capitalize]{cleveref}
\crefname{section}{Sec.}{Secs.}
\Crefname{section}{Section}{Sections}
\Crefname{table}{Table}{Tables}
\crefname{table}{Tab.}{Tabs.}

\def\cvprPaperID{2233} 
\def\confName{CVPR}
\def\confYear{2023}

\begin{document}

\title{TryOnDiffusion:  A Tale of Two UNets}

\author{
Luyang Zhu\textsuperscript{1,2}\footnotemark\qquad\qquad 
Dawei Yang\textsuperscript{2}\qquad 
Tyler Zhu\textsuperscript{2}\qquad
Fitsum Reda\textsuperscript{2}\qquad
William Chan\textsuperscript{2}\qquad \\
Chitwan Saharia\textsuperscript{2}\qquad
Mohammad Norouzi\textsuperscript{2}\qquad
Ira Kemelmacher-Shlizerman\textsuperscript{1,2} \\
\textsuperscript{1}University of Washington\qquad 
\textsuperscript{2}Google Research
}

\twocolumn[{%
\renewcommand\twocolumn[1][]{#1}%
\maketitle
\begin{center}
    \centering
    \includegraphics[width=.97\textwidth]{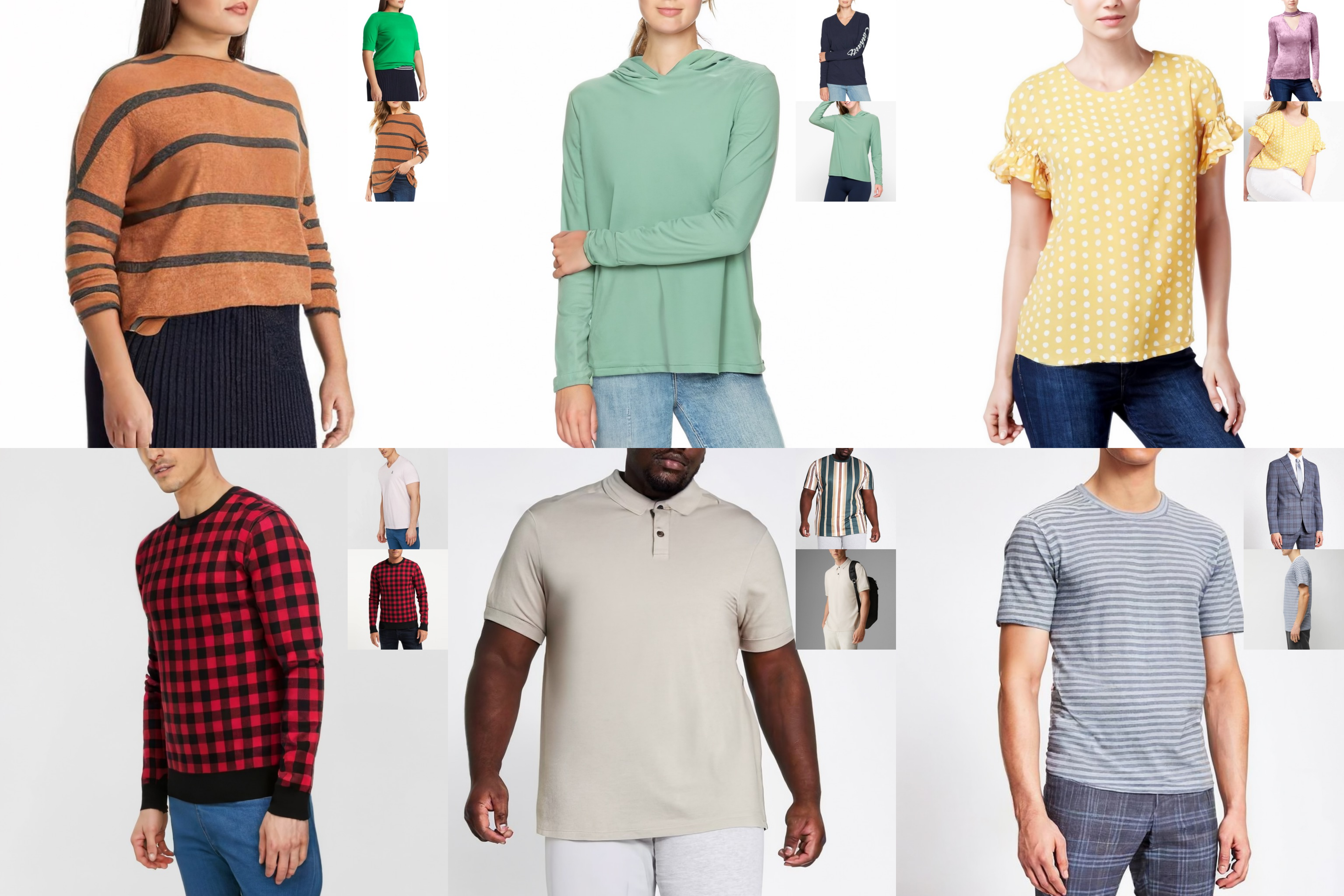}
    \captionof{figure}{\small TryOnDiffusion generates apparel try-on results with a significant body shape and pose modification, while preserving garment details at $1024 \mathord\times\mathord1024$ resolution. Input images (target person and garment worn by another person) are shown in the corner of the results.
    }
    \label{fig:teaser}
\end{center}%
}]

\footnotetext[1]{Work done while author was an intern at Google.}

\maketitle


\begin{abstract}
Given two images depicting a person and a garment worn by another person, our goal is to generate a visualization of how the garment might look on the input person. A key challenge is to synthesize a photorealistic detail-preserving visualization of the garment, while warping the garment to accommodate a significant body pose and shape change across the subjects. Previous methods either focus on garment detail preservation without effective pose and shape variation, or allow try-on with the desired shape and pose but lack garment details. In this paper, we propose a diffusion-based architecture that unifies two UNets (referred to as {\unet}), which allows us to preserve garment details {and} warp the garment for significant pose and body change in a single network. The key ideas behind {\unet} include: 1) garment is warped implicitly via a cross attention mechanism, 2) garment warp and person blend happen as part of a unified process as opposed to a sequence of two separate tasks.
Experimental results indicate that TryOnDiffusion achieves state-of-the-art performance both qualitatively and quantitatively.%
\vspace{-4mm}
\end{abstract}

\begin{figure*}
\begin{center}
  \includegraphics[width=1.0\linewidth]{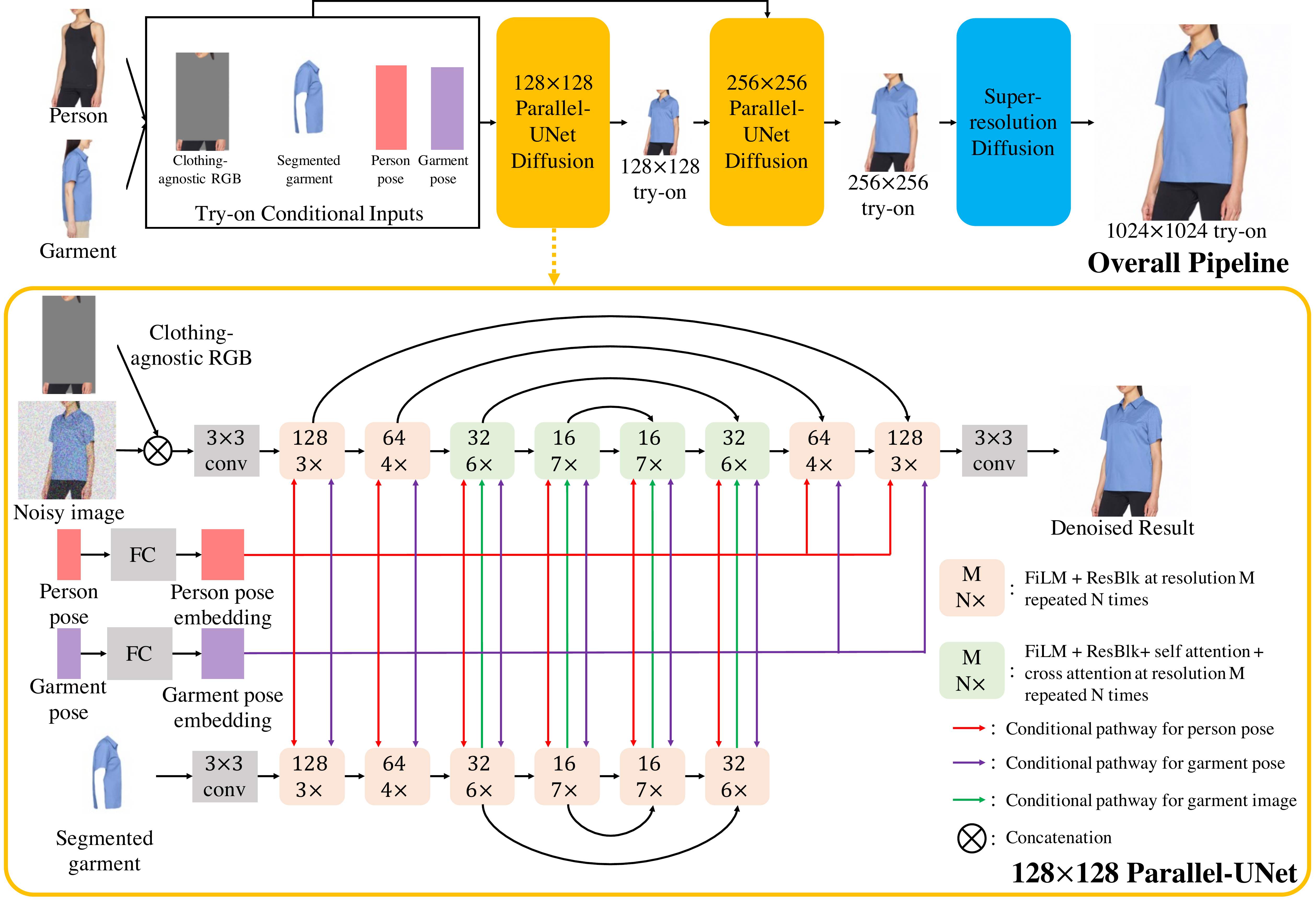}
\end{center}
\vspace{-8mm}
  \caption{Overall pipeline (top): During preprocessing step, the target person is segmented out of the person image creating ``clothing agnostic RGB" image,  the target garment is segmented out of the garment image, and pose is computed for both person and garment images. These inputs are taken into $128 \mathord\times\mathord128$ \unet (key contribution) to create the $128\times 128$ try-on image which is further sent as input to the 
  $256 \mathord\times\mathord256$ \unet together with the try-on conditional inputs. Output from $256 \mathord\times\mathord256$ \unet is sent to standard super resolution diffusion to create the $1024 \mathord\times\mathord1024$ image.   The architecture of $128 \mathord\times\mathord128$ \unet is visualized at the bottom, see text for details. The $256 \mathord\times\mathord256$ \unet is similar to the 128 one, and  provided in supplementary for completeness.
  }
\vspace{-3mm}
\label{fig:pipeline}
\end{figure*}

\section{Introduction}

\label{sec::intro}
Virtual apparel try-on aims to visualize how a garment might look on a person based on an image of the person and an image of the garment.
Virtual try-on has the potential to enhance the online shopping experience, but most try-on methods only perform well when
body pose and shape variation is small. 
A key open problem is the non-rigid warping of a garment to fit a target body shape, while not introducing distortions in garment patterns and texture~\cite{han2018viton,wang2018toward,choi2021viton}. 

When pose or body shape vary significantly, garments need to warp in a way that wrinkles are created or flattened according to the new shape or occlusions. Related works~\cite{choi2021viton,lee2022hrviton,walmart} have been approaching the warping problem via first estimating pixel displacements, \eg, optical flow, followed by pixel warping, and postprocessing with perceptual loss when blending with the target person. Fundamentally, however, the sequence of finding displacements, warping, and blending often creates artifacts, since occluded parts and shape deformations are challenging to model accurately with pixel displacements. It is also challenging to remove those artifacts later in the blending stage even if it is done with a powerful generative model. As an alternative, TryOnGAN~\cite{lewis2021tryongan} showed how to warp without estimating displacements, via a conditional  StyleGAN2~\cite{karras2020analyzing} network and optimizing in generated latent space. While the generated results were of impressive quality, outputs often lose details especially for highly patterned garments due to the low representation power of the latent space. 

In this paper, we present TryOnDiffusion that can handle large occlusions, pose changes, and body shape changes, while preserving garment details at $1024 \mathord\times\mathord1024$ resolution. TryOnDiffusion takes as input two images: a target person image, and an image of a garment worn by another person. It synthesizes as output the target person wearing the garment. The garment might be partially occluded by body parts or other garments, and requires significant deformation. Our method is trained on 4 Million image pairs. Each pair has the same person wearing the same garment but appears in different poses. 

TryOnDiffusion is based on our novel architecture called \unet consisting of two sub-UNets communicating through cross attentions~\cite{vaswani2017attention}.
Our two key design elements are implicit warping and combination of warp and blend (of target person and garment) in a single pass rather than in a sequential fashion.
Implicit warping between the target person and the source garment is achieved via cross attention over their features at multiple pyramid levels which allows to establish long range correspondence.
Long range correspondence performs well, especially under heavy occlusion and extreme pose differences. Furthermore, using the same network to perform warping and blending allows the two processes to exchange information at the feature level rather than at the color pixel level which proves to be essential in perceptual loss and style loss~\cite{johnson2016perceptual,reda2022film}. We demonstrate the performance of these design choices in Sec.~\ref{sec::experiments}. 

To generate high quality results at $1024 \mathord\times\mathord1024$ resolution, we follow Imagen~\cite{saharia2022photorealistic} and create cascaded diffusion models. Specifically, \unet{} based diffusion is used for $128 \mathord\times\mathord128$ and $256 \mathord\times\mathord256$ resolutions. The $256 \mathord\times\mathord256$ result is then fed to a super-resolution diffusion network to create the final $1024 \mathord\times\mathord1024$ image.  

In summary, the main contributions of our work are: 1) try-on synthesis at $1024 \mathord\times\mathord1024$ resolution for a variety of complex body poses, allowing for diverse body shapes, while preserving garment details (including patterns, text, labels, etc.),
2) a novel architecture called \unet{}, which can warp the garment implicitly with cross attention, in addition to warping and blending in a single network pass.
We evaluated TryOnDiffusion quantitatively and qualitatively, compared to recent state-of-the-art methods, and performed an extensive user study. The user study was done by $15$ non-experts,  ranking more than 2K distinct random samples. The study showed that our results were chosen as the best $92.72\%$ of the time compared to three recent state-of-the-art methods. 


\section{Related Work}
\label{sec::related}
\mypar{Image-Based Virtual Try-On.} Given a pair of images (target person, source garment), image-based virtual try-on methods generate the look of the target person wearing the source garment. Most of these methods~\cite{han2018viton,wang2018toward,yu2019vtnfp,issenhuth2020not,yang2020towards,choi2021viton,ge2021parser,dong2022dressing,He_2022_CVPR,Yang_2022_CVPR,lee2022hrviton,bai2022single,men2020controllable,zhang2021pise,ren2022neural} decompose the try-on task into two stages: a warping model and a blending model. The seminal work VITON~\cite{han2018viton} proposes a coarse-to-fine pipeline guided by the thin-plate-spline (TPS) warping of source garments. 
ClothFlow~\cite{han2019clothflow} directly estimates flow fields with a neural network instead of TPS for better garment warping.
VITON-HD~\cite{choi2021viton} introduces alignment-aware generator to increase the try-on resolution from $256 \mathord\times\mathord192$ to $1024 \mathord\times\mathord768$. HR-VITON~\cite{lee2022hrviton} further improves VITON-HD by predicting segmentation and flow simultaneously. SDAFN~\cite{bai2022single} predicts multiple flow fields for both the garment and the person, and combines warped features through deformable attention~\cite{zhu2020deformable} to improve quality.

Despite great progress, these methods still suffer from  misalignment brought by  explicit flow estimation and warping. TryOnGAN~\cite{lewis2021tryongan} tackles this issue by training a pose-conditioned StyleGAN2~\cite{karras2020analyzing} on unpaired fashion images and running optimization in the latent space to achieve try-on. By optimizing the latent space, however, it loses garment details that are less represented by the latent space. This becomes evident when garments have a pattern or details like pockets, or special sleeves. 

We propose a novel architecture which  performs implicit warping (without computing flow) and blending in a single network pass. Experiments show that our method can preserve details of the garment even under heavy occlusions and various body poses and shapes.

\mypar{Diffusion Models.} Diffusion models~\cite{sohl2015deep,song2019generative,ho2020denoising} have recently emerged as the most powerful family of generative models. Unlike GANs~\cite{goodfellow2020generative,brock2018large}, diffusion models have better training stability and mode coverage. They have achieved state-of-the-art results on various image generation tasks, such as super-resolution~\cite{saharia2022image}, colorization~\cite{saharia2022palette}, novel-view synthesis~\cite{watson2022novel} and text-to-image generation~\cite{saharia2022photorealistic,ramesh2022hierarchical,rombach2022high,ruiz2022dreambooth}. Although being successful, state-of-the-art diffusion models utilize a traditional UNet architecture~\cite{ronneberger2015u,ho2020denoising} with channel-wise concatenation~\cite{saharia2022image,saharia2022palette} for image conditioning. The channel-wise concatenation works well for image-to-image translation problems where input and output pixels are perfectly aligned (\eg, super-resolution, inpainting and colorization). However, it is not directly applicable to our task as try-on involves highly non-linear transformations like garment warping. To solve this challenge, we propose \unet{} architecture tailored to try-on, where the garment is warped implicitly via cross attentions.


\section{Method}
\label{sec::method}
Fig.~\ref{fig:pipeline} provides an overview of our method for virtual try-on. Given an image $I_{p}$ of person $p$ and an image $I_{g}$ of a different person in garment $g$, our approach generates  try-on result $I_{\text{tr}}$ of person $p$ wearing garment $g$.  Our  method is trained on paired data where $I_{p}$ and $I_{g}$ are images of the same person wearing the same garment but in two different poses. During inference, $I_{p}$ and $I_{g}$ are set to images of two different people wearing different garments in different poses. We begin by describing our preprocessing steps, and a brief paragraph on diffusion models. Then we describe in subsections our  contributions and design choices.

\label{subsec::tryon_cond_inputs}
\mypar{Preprocessing of inputs.} We first predict  human parsing map ($S_{p},S_{g}$) and 2D pose keypoints ($J_{p},J_{g}$) for both person and garment images using off-the-shelf methods~\cite{gong2019graphonomy,papandreou2017towards}. For garment image, we further segment out the garment $I_{c}$ using the parsing map. For person image, we generate  clothing-agnostic RGB image $I_{a}$ which removes the original clothing but retains the person identity. Note that clothing-agnostic RGB described in VITON-HD~\cite{choi2021viton} leaks information of the original garment for challenging human poses and loose garments. We thus adopt a more aggressive way to remove the garment information. Specifically, we first mask out the whole bounding box area of the foreground person, and then copy-paste the head, hands and lower body part on top of it. We use $S_{p}$ and $J_{p}$ to extract the non-garment body parts. We also normalize pose keypoints to the range of $[0,1]$ before inputting them to our networks. Our try-on conditional inputs are denoted as $\mathbf{c}_\text{tryon} = (I_{a}, J_{p}, I_{c}, J_{g})$.

\label{sec::preliminaries}
\mypar{Brief overview of diffusion models.} Diffusion models \cite{sohl2015deep,ho2020denoising} are a class of generative models that learn the target distribution through an iterative denoising process. They consist of a Markovian forward process that gradually corrupts the data sample $\mathbf{x}$ into the Gaussian noise $\mathbf{z}_{T}$, and a learnable reverse process that converts $\mathbf{z}_{T}$ back to $\mathbf{x}$ iteratively. Diffusion models can be conditioned on various signals such as class labels, texts or images. A conditional diffusion model $\hat{\mathbf{x}}_{\theta}$ can be trained with a weighted denoising score matching objective:
\begin{equation}
\label{eq:diffusion}
    \mathbb{E}_{\mathbf{x},\mathbf{c},\boldsymbol{\epsilon},t}[{w_t \|\hat{\mathbf{x}}_{\theta}(\alpha_t \mathbf{x} + \sigma_t \boldsymbol{\epsilon}, \mathbf{c}) - \mathbf{x} \|^2_2}]
\end{equation}
where $\mathbf{x}$ is the target data sample, $\mathbf{c}$ is the conditional input, $\boldsymbol{\epsilon} \sim \mathcal{N}(\mathbf{0}, \mathbf{I})$ is the noise term. $\alpha_t, \sigma_t, w_t$ are functions of the timestep $t$ that affect sample quality. In practice, $\hat{\mathbf{x}}_{\theta}$ is reparameterized as $\hat{\boldsymbol{\epsilon}}_{\theta}$ to predict the noise that corrupts $\mathbf{x}$ into $\mathbf{z}_t :=\alpha_t \mathbf{x} + \sigma_t \boldsymbol{\epsilon}$. At inference time, data samples can be generated from Gaussian noise $\mathbf{z}_{T} \sim \mathcal{N}(\mathbf{0}, \mathbf{I})$ using samplers like DDPM~\cite{ho2020denoising} or DDIM~\cite{song2020denoising}.

\subsection{Cascaded Diffusion Models for Try-On}
\label{subsec::cascade_diffusion}
Our cascaded diffusion models consist of one base diffusion model and two super-resolution (SR) diffusion models.

The base diffusion model is parameterized as a $128\mathord\times\mathord128$ \unet{} (see Fig.~\ref{fig:pipeline} bottom). It predicts the $128\mathord\times\mathord128$ try-on result $I_{\text{tr}}^{128}$, taking in the try-on conditional inputs $\mathbf{c}_\text{tryon}$. Since $I_{a}$ and $I_{c}$ can be noisy due to inaccurate human parsing and pose estimations, we apply  noise conditioning augmentation~\cite{ho2022cascaded} to them. Specifically, random Gaussian noise is added to $I_{a}$ and $I_{c}$ before any other processing. The levels of noise augmentation are also treated as conditional inputs following~\cite{ho2022cascaded}.

The $128\mathord\times\mathord128\mathord\rightarrow\mathord256\mathord\times\mathord256$ SR diffusion model is parameterized as a $256\mathord\times\mathord256$ \unet{}. It generates the $256\mathord\times\mathord256$ try-on result $I_{\text{tr}}^{256}$ by conditioning on both the $128\mathord\times\mathord128$ try-on result $I_{\text{tr}}^{128}$ and the try-on conditional inputs $\mathbf{c}_\text{tryon}$ at $256\mathord\times\mathord256$ resolution. $I_{\text{tr}}^{128}$ is directly downsampled from the ground-truth during training. At test time, it is set to the prediction from the base diffusion model. Noise conditioning augmentation is applied to all conditional input images at this stage, including $I_{\text{tr}}^{128}$, $I_{a}$ and $I_{c}$.

The $256\mathord\times\mathord256\mathord\rightarrow\mathord1024\mathord\times\mathord1024$ SR diffusion model is parameterized as Efficient-UNet introduced by Imagen~\cite{saharia2022photorealistic}. This stage is a pure super-resolution model, with no try-on conditioning. For training, random $256\mathord\times\mathord256$ crops, from $1024\mathord\times\mathord1024$, serve as the ground-truth, and the input is set to $64\mathord\times\mathord64$ images downsampled from the crops. During inference, the model takes as input $256\mathord\times\mathord256$ try-on result from previous Parallel-UNet model and synthesizes the final try-on result $I_{\text{tr}}$ at $1024\mathord\times\mathord1024$ resolution. To facilitate this setting, we make the network fully convolutional by removing all attention layers. Like the two previous models, noise conditioning augmentation is applied to the conditional input image.

\subsection{\unet}
\label{subsec::parallel_unet}
The $128\mathord\times\mathord128$ \unet can be represented as 
\begin{equation}
\label{eq:classifier-free guidance}
    \boldsymbol{\epsilon}_t = \boldsymbol{\epsilon}_\theta(\mathbf{z}_t, t, \mathbf{c}_\text{tryon}, \mathbf{t}_{\text{na}})
\end{equation}

where $t$ is the diffusion timestep, $\mathbf{z}_t$ is the noisy image corrupted from the ground-truth at timestep $t$, $\mathbf{c}_\text{tryon}$ is the try-on conditional inputs, $\mathbf{t}_{\text{na}}$ is the set of noise augmentation levels for different conditional images, and $\boldsymbol{\epsilon}_t$ is predicted noise that can be used to recover the ground-truth from $\mathbf{z}_t$. The $256\mathord\times\mathord256$ \unet takes in the try-on result $I_{\text{tr}}^{128}$ as input, in addition to the try-on conditional inputs  $\mathbf{c}_\text{tryon}$ at $256\mathord\times\mathord256$ resolution.
Next, we describe two key design elements of \unet.

\mypar{Implicit warping.} The first question is: how can we implement implicit warping in the neural network? 
One natural solution is to use a traditional UNet~\cite{ronneberger2015u,ho2020denoising} and concatenate the segmented garment $I_{c}$ and the noisy image $\mathbf{z}_t$ along the channel dimension.
However, channel-wise concatenation\cite{saharia2022image,saharia2022palette} can not handle complex transformations such as garment warping (see Sec.~\ref{subsec::ablative}).
This is because the computational primitives of the traditional UNet are spatial convolutions and spatial self attention, and these primitives have strong pixel-wise structural bias.
To solve this challenge, we propose to achieve implicit warping using cross attention mechanism between our streams of information ($I_{c}$ and $\mathbf{z}_t$).
The cross attention is based on the scaled dot-product attention introduced by \cite{vaswani2017attention}:
\begin{equation}
\label{eq:attention}
    \text{Attention}(Q,K,V)=\text{softmax}(\frac{QK^T}{\sqrt{d}})V
\end{equation}
where $Q \in \mathbb{R}^{M\times d}, K\in \mathbb{R}^{N\times d}, V\in \mathbb{R}^{N\times d}$ are stacked vectors of query, key and value, $M$ is the number of query vectors, $N$ is the number of key and value vectors and $d$ is the dimension of the vector. In our case, the query and key-value pairs come from different inputs. Specifically, $Q$ is the flattened features of $\mathbf{z}_t$ and $K, V$ are the flattened features of $I_{c}$. The attention map $\frac{QK^T}{\sqrt{d_k}}$ computed through dot-product tells us the similarity between the target person and the source garment, providing a learnable way to represent correspondence for the try-on task. We also make the cross attention multi-head, allowing the model to learn from different representation subspaces.

\mypar{Combining warp and blend in a single pass.}
Instead of warping the garment to the target body and then blending with the target person as done by prior works, we combine the two operations into a single pass. As shown in Fig.~\ref{fig:pipeline}, we achieve it via two UNets that handle the garment and the person respectively.

The person-UNet takes the clothing-agnostic RGB $I_{a}$ and the noisy image $\mathbf{z}_t$ as input. Since $I_{a}$ and $\mathbf{z}_t$ are pixel-wise aligned, we directly concatenate them along the channel dimension at the beginning of UNet processing.

The garment-UNet takes the segmented garment image $I_{c}$ as input. The garment features are fused to the target image via cross attentions defined above.
To save model parameters, we early stop the garment-UNet after the $32\mathord\times\mathord32$ upsampling block, where the final cross attention module in person-UNet is done.

The person and garment poses are necessary for guiding the warp and blend process. They are first fed into the linear layers to compute pose embeddings separately. The pose embeddings are then fused to the person-UNet through the attention mechanism, which is implemented by concatenating pose embeddings to the key-value pairs of each self attention layer~\cite{saharia2022photorealistic}. Besides, pose embeddings are reduced along the keypoints dimension using CLIP-style 1D attention pooling~\cite{radford2021learning}, and summed with the positional encoding of diffusion timestep $t$ and noise augmentation levels $\mathbf{t}_{\text{na}}$. The resulting 1D embedding is used to modulate features for both UNets using FiLM~\cite{dumoulin2018feature} across all scales.


\begin{table}[tb]
\centering
\begin{tabular}{ |c|cc|cc|  }
\hline
\multicolumn{1}{|c}{Test datasets} & \multicolumn{2}{|c}{Ours } & \multicolumn{2}{|c|}{VITON-HD}\\
\hline
Methods & FID $\downarrow$ & KID $\downarrow$ & FID $\downarrow$ & KID $\downarrow$\\ 
\hline 
TryOnGAN~\cite{lewis2021tryongan} & 24.577   &  16.024 & 30.202 & 18.586 \\
SDAFN~\cite{bai2022single} &  18.466  & 10.877 & 33.511 & 20.929  \\
HR-VITON~\cite{lee2022hrviton} & 18.705   &  9.200 & 30.458 & 17.257 \\
Ours & \textbf{13.447} & \textbf{6.964} & \textbf{23.352} & \textbf{10.838} \\
 \hline
\end{tabular}
\vspace{-3mm}
\caption{Quantitative comparison to 3 baselines. We compute FID and KID on our 6K test set and VITON-HD's unpaired test set. The KID is scaled by 1000 following \cite{karras2020training}.}
\label{table:quantitative}
\vspace{-5mm}
\end{table}

\begin{table}[htb]
\centering
\begin{tabular}{ |c|c|c|}
\hline
Methods & Random & Challenging\\ 
\hline 
TryOnGAN~\cite{lewis2021tryongan}&  1.75\%  & 0.45\% \\
SDAFN~\cite{bai2022single}  &  2.42\%  &  2.20\% \\
HR-VITON~\cite{lee2022hrviton} &  2.92\%  & 1.30\%\\
Ours & \textbf{92.72\%} & \textbf{95.80\%} \\
Hard to tell &  0.18\% & 0.25\% \\
\hline
\end{tabular}
\vspace{-3mm}
\caption{Two user studies. ``Random": 2804 random input pairs (out of 6K) were rated by 15 non-experts asked to select the best result or choose ``hard to tell".
``Challenging": 2K pairs with challenging body poses were selected out of 6K and rated in same fashion. Our method significantly outperforms others in both studies.}
\label{table:user_study}
\vspace{-5mm}
\end{table}

\begin{figure*}[htb]
\begin{center}
   \includegraphics[width=1.0\linewidth]{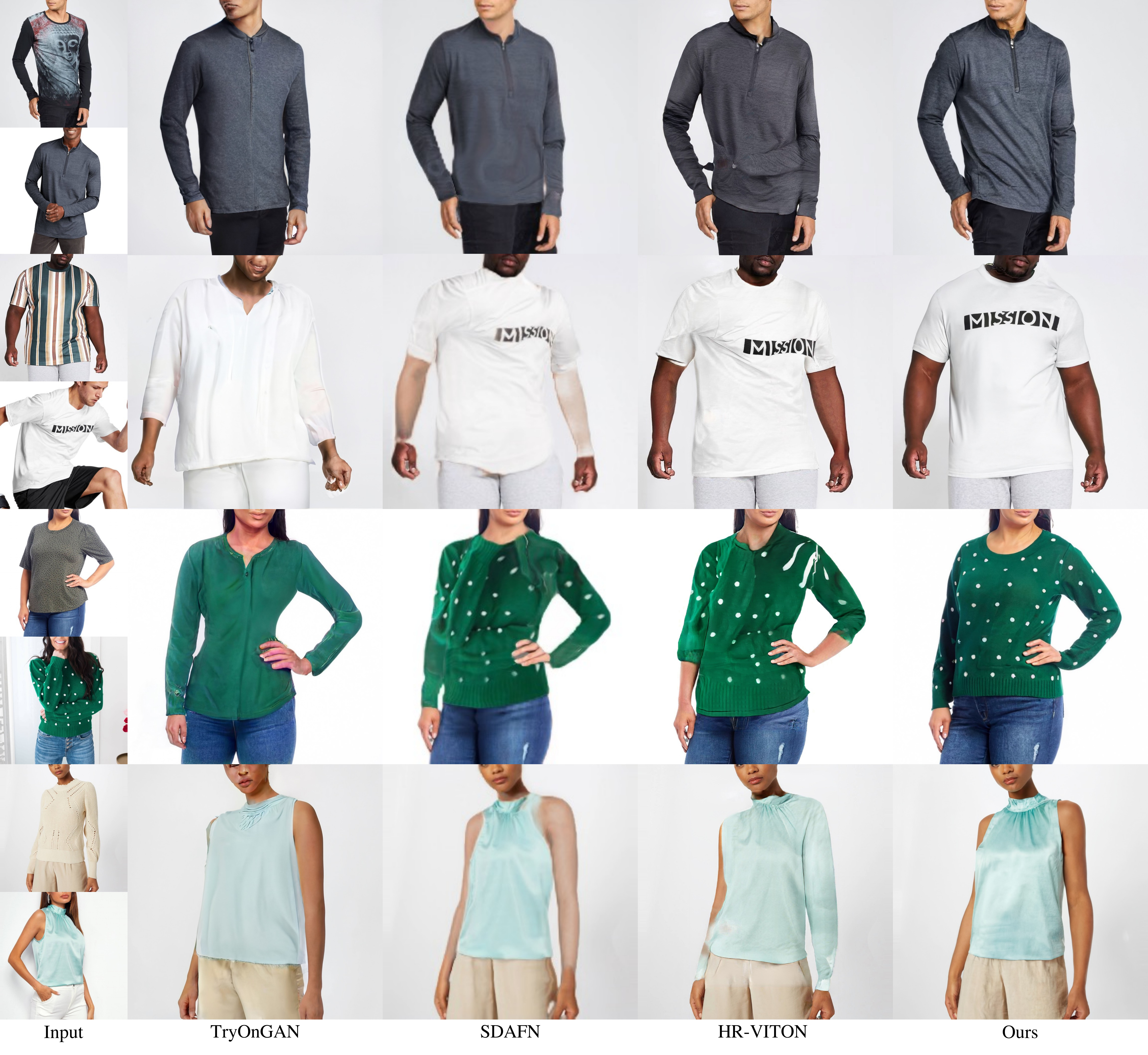}
\end{center}
\vspace{-7mm}
\caption{Comparison with TryOnGAN~\cite{lewis2021tryongan}, SDAFN~\cite{bai2022single} and HR-VITON~\cite{lee2022hrviton}. First column shows the input (person, garment) pairs. TryOnDiffusion warps well garment details including  text and  geometric patterns even under extreme body pose and shape  changes.}
\label{fig:sota_comparison}
\vspace{-3mm}
\end{figure*}

\begin{figure*}[htb]
\begin{center}
   \includegraphics[width=0.97\linewidth]{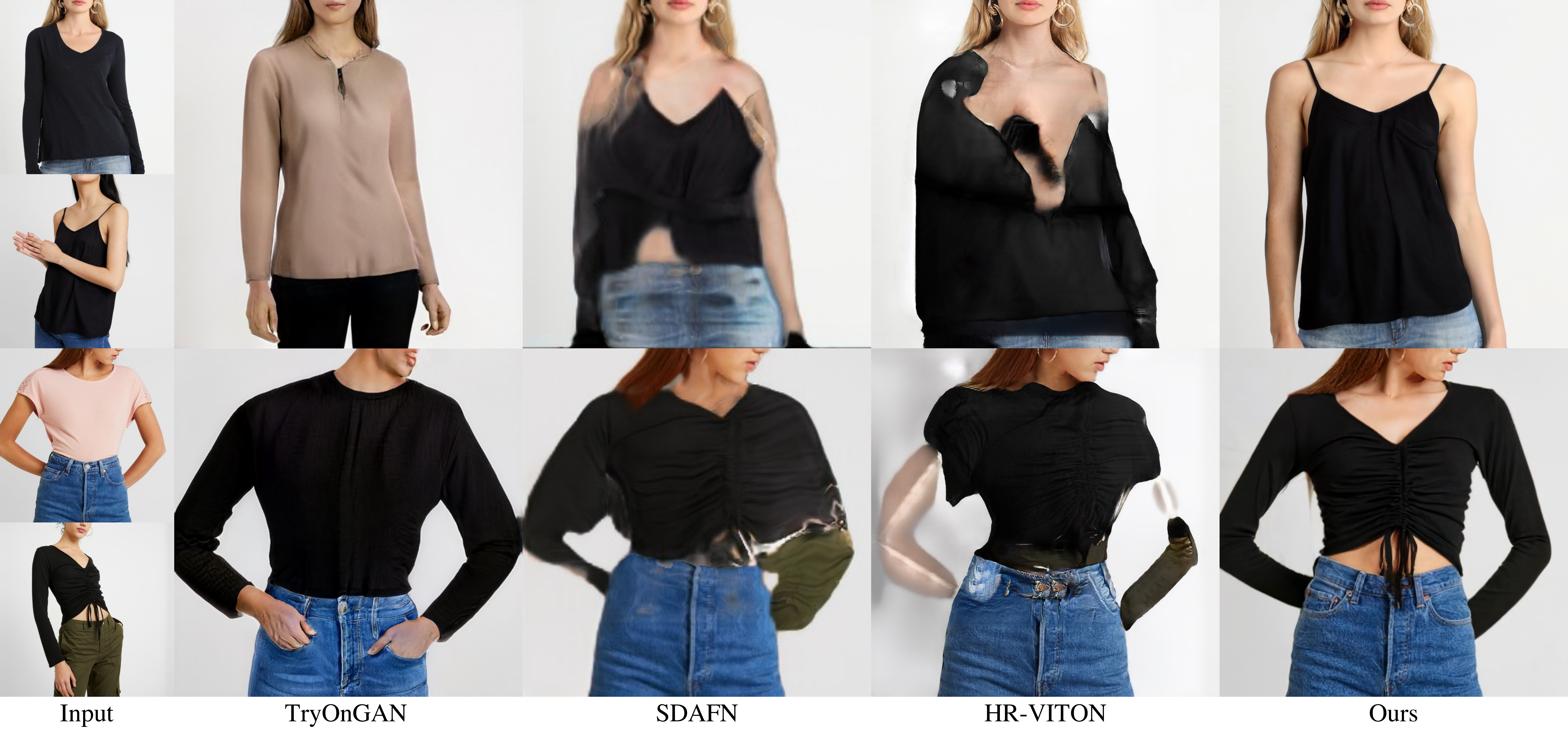}
\end{center}
\vspace{-7mm}
\caption{Comparison with state-of-the-art methods on VITON-HD dataset~\cite{choi2021viton}. All methods were trained on the same 4M dataset and tested on VITON-HD.}
\label{fig:vitonhd_dataset}
\vspace{-3mm}
\end{figure*}

\begin{figure*}[htb]
\begin{center}
  \includegraphics[width=1.0\linewidth]{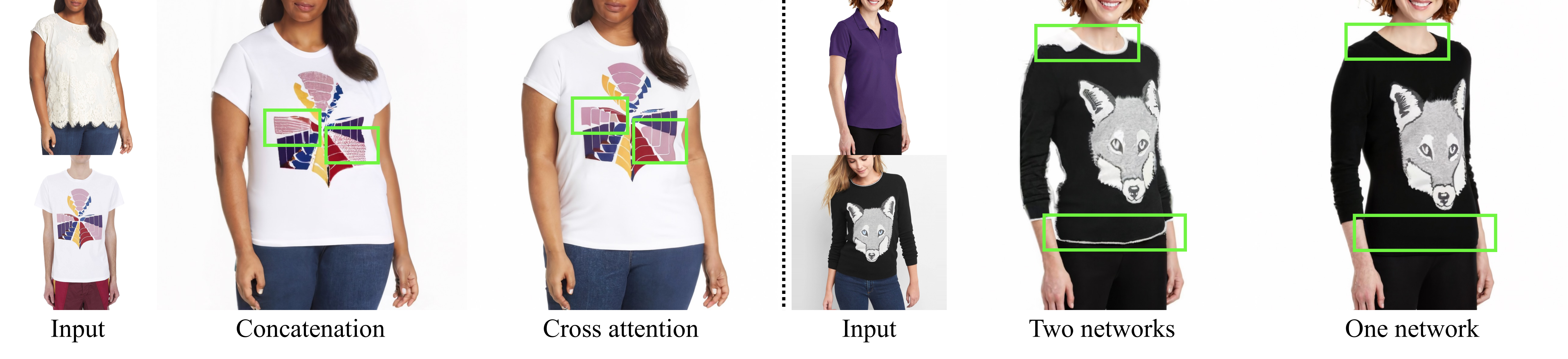}
\end{center}
\vspace{-7mm}
\caption{Qualitative results for ablation studies. Left: cross attention versus concatenation for implicit warping. Right: One network versus two networks for warping and blending. Zoom in to see differences highlighted by green boxes.}
\label{fig:ablate}
\vspace{-2mm}

\end{figure*}

\begin{figure*}[htb]
\begin{center}
  \includegraphics[width=1.0\linewidth]{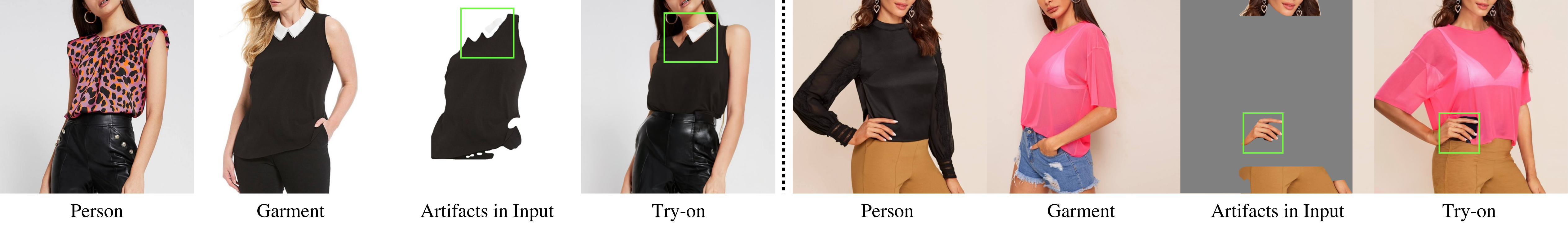}
\end{center}
\vspace{-7mm}
\caption{Failures happen due to erroneous garment segmentation (left) or  garment leaks into the Clothing-agnostic RGB image (right). 
}
\label{fig:failure_case}
\vspace{-2mm}
\end{figure*}

\begin{figure*}[htb]
\begin{center}
   \includegraphics[width=1.0\linewidth]{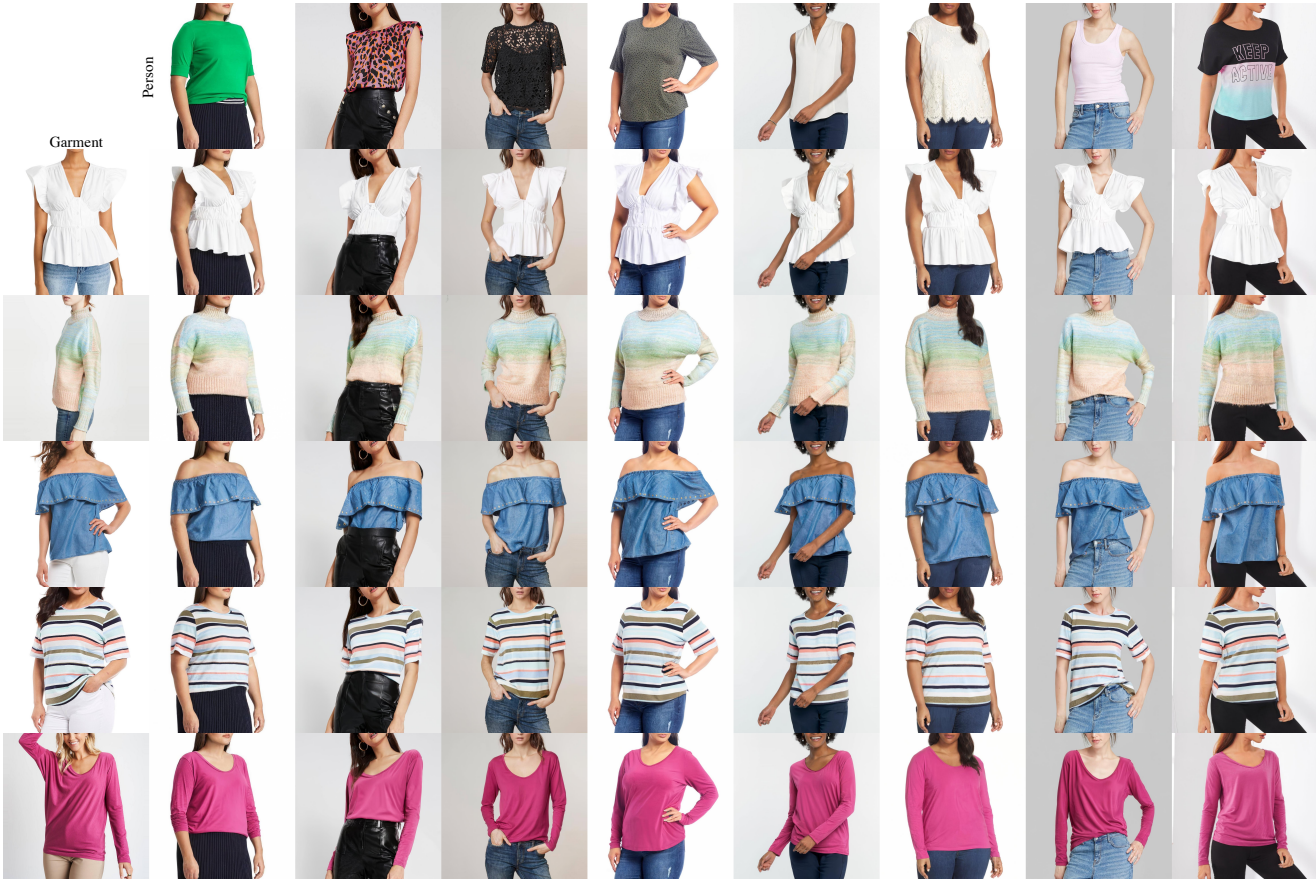}
\end{center}
\vspace{-5mm}
\caption{TryOnDiffusion on eight target people (columns) dressed by five garments (rows). Zoom in to see details.}
\label{fig:venti_tryon}
\vspace{-6mm}
\end{figure*}

\section{Experiments}
\label{sec::experiments}

\label{subsec::exp_setup}
\mypar{Datasets.} We collect a paired training dataset of 4 Million samples. Each sample consists of two images of the same person wearing the same garment in two different poses. For test, we collect 6K unpaired samples that are never seen during training. Each test sample includes two images of \textit{different} people wearing \textit{different} garments under \textit{different} poses. Both training and test images are cropped and resized to $1024 \mathord\times\mathord1024$ based on detected 2D human poses. Our dataset includes both men and women captured  in different poses, with different body shapes,  skin tones, and wearing a wide variety of garments with diverse texture patterns. 
In addition, we also provide results on the VITON-HD dataset~\cite{choi2021viton}.

    \mypar{Implementation details.} All three models are trained with batch size 256 for 500K iterations using the Adam optimizer~\cite{kingma2014adam}. The learning rate linearly increases from $0$ to $10^{-4}$ for the first 10K iterations and is kept constant afterwards. We follow classifier-free guidance~\cite{ho2022classifier} and train our models with conditioning dropout: conditional inputs are set to 0 for $10\%$ of training time. All of our test results are generated with the following schedule: The base diffusion model is sampled for 256 steps using DDPM; The $128 \mathord\times\mathord128 \mathord\rightarrow\mathord 256 \mathord\times\mathord256$ SR diffusion model is sampled for 128 steps using DDPM; The final $256 \mathord\times\mathord256 \mathord\rightarrow\mathord 1024 \mathord\times\mathord1024$ SR diffusion model is sampled for 32 steps using DDIM. The guidance weight is set to $2$ for all three stages. During training, levels of noise conditioning augmentation are sampled from uniform distribution $\mathcal{U}([0,1])$. At inference time, they are set to constant values based on  grid search, following~\cite{saharia2022photorealistic}.

\label{subsec::compare_sota}
\mypar{Comparison with other methods.} We compare our approach to three methods: TryOnGAN~\cite{lewis2021tryongan}, SDAFN~\cite{bai2022single} and HR-VITON~\cite{lee2022hrviton}. For fair comparison, we re-train all three methods on our 4 Million  samples until convergence. Without re-training, the results of these methods are worse. Released checkpoints of SDAFN and HR-VITON also require layflat garment as input, which is not applicable to our setting. The resolutions of the related methods vary, and we present each method's results in their native resolution: SDAFN's at $256 \mathord\times\mathord256$, TryOnGAN's at $512\times 512$ and HR-VITON at $1024\times 1024$.

\mypar{Quantitative comparison.} Table~\ref{table:quantitative} provides  comparisons with two metrics.  Since our test dataset is unpaired, we compute Frechet Inception Distance (FID)~\cite{heusel2017gans} and Kernel Inception Distance (KID)~\cite{binkowski2018demystifying} as evaluation metrics.  We computed those metrics on both test datasets (our 6K, and VITON-HD) and observe a significantly better performance with our method. 

\mypar{User study.} We ran two user studies to objectively evaluate our methods compared to others at scale. The results are reported in Table~\ref{table:user_study}. In first study (named ``random"), we randomly selected $2804$ input pairs out of the 6K test set, ran all four methods on those pairs, and presented to raters. 15 non-expert raters (on crowdsource platform) have been asked to select the best result out of four or choose ``hard to tell" option. Our method was selected as best for 92.72\% of the inputs.  In a second study (named ``challenging"), we performed the same setup but chose 2K input pairs (out of 6K) with more challenging poses. The raters selected our method as best for 95.8\% of the inputs.

\mypar{Qualitative comparison.} In Figures~\ref{fig:sota_comparison} and \ref{fig:vitonhd_dataset}, we provide visual comparisons to all baselines on two test datasets (our 6K, and VITON-HD). Note that many of the chosen input pairs have quite different body poses, shapes and complex garment materials--all limitations of most previous methods--thus we don't expect them to perform well but present here to show the strength of our method.  Specifically,  we observe that TryOnGAN struggles to retain the texture pattern of the garments while SDAFN and HR-VITON introduce warping artifacts in the try-on results. In contrast, our approach preserves fine details of the source garment and seamlessly blends the garment with the person even if the poses are hard or materials are complex (Fig.~\ref{fig:sota_comparison}, row 4).  Note also how TryOnDiffusion generates realistic garment wrinkles corresponding to the new body poses (Fig.~\ref{fig:sota_comparison}, row 1). We show easier poses in the supplementary (in addition to more results) to provide a fair comparison to other methods.

\label{subsec::ablative}

\mypar{Ablation 1: Cross attention vs concatenation for implicit warping.}  The implementation of cross attention is detailed in Sec.~\ref{subsec::parallel_unet}. For concatenation, we discard the garment-UNet, directly concatenate the segmented garment $I_{c}$ to the noisy image $\mathbf{z}_t$, and drop cross attention modules in the person-UNet. We apply these changes to each \unet, and keep the final SR diffusion model same. Fig.~\ref{fig:ablate} shows that cross attention is better at preserving garment details under significant body pose and shape changes.

\mypar{Ablation 2: Combining warp and blend vs sequencing two tasks.}  Our method combines both steps in one network pass as described in Sec.~\ref{subsec::parallel_unet}. For the ablated version, we train two base diffusion models while SR diffusion models are intact. The first base diffusion model handles the warping task. It takes as input the segmented garment $I_{c}$, the person pose $J_{p}$ and the garment pose $J_{g}$, and  predicts the warped garment $I_{wc}$. The second base diffusion model performs the blending task, whose inputs are the warped garment $I_{wc}$,  clothing-agnostic RGB $I_{a}$,  person pose $J_{p}$ and  garment pose $J_{g}$. The output is the try-on result $I_{\text{tr}}^{128}$ at $128\mathord\times\mathord128$ resolution. The conditioning for $(I_{c}, I_{a}, J_{p}, J_{g})$ is kept unchanged. $I_{wc}$ in the second base diffusion model is processed by a garment-UNet, which is the same as $I_{c}$. Fig.~\ref{fig:ablate} visualizes the results of both methods. We can see that sequencing warp and blend causes artifacts near the garment boundary, while a single network can blend the target person and the source garment nicely.

\label{subsec::falure_cases}
\mypar{Limitations.} First, our method  exhibits garment leaking artifacts in case of errors in segmentation maps and pose estimations during preprocessing. Fortunately, those ~\cite{papandreou2017towards,gong2019graphonomy} became quite accurate in recent years and this does not happen often. Second, representing identity via clothing-agnostic RGB  is not ideal, since sometimes it may preserve only part of the identity, \eg, tatooes won't be visible in this representation, or specific muscle structure.  Third, our train and test datasets have mostly clean uniform background so it is unknown how the method performs with more complex backgrounds. Finally, this work focused on  upper body clothing and we have not experimented with full body try-on, which is left for future work. Fig.~\ref{fig:failure_case} demonstrates failure cases.

Finally, Fig.~\ref{fig:venti_tryon} shows TryOnDiffusion results on variety of people and garments.  Please refer to  supplementary material for more  results. 


\vspace{-3mm}
\section{Summary and Future Work}
We presented a method that allows to synthesize try-on given an image of a person and an image of a garment. Our results are overwhelmingly better than state-of-the-art, both in the quality of the warp to new body shapes and poses, and in the preservation of the garment.
Our novel architecture \unet, where two UNets are trained in parallel and one UNet sends information to the other via cross attentions, turned out to create state-of-the-art results. 
In addition to the exciting progress for the specific application of virtual try-on, we believe this architecture is going to be impactful for the general case of image editing, which we are excited to explore in the future. Finally, we believe that the architecture could also be extended to videos, which we also plan to pursue in the future. 
\newpage

{\small
\bibliographystyle{ieee_fullname}
\bibliography{egbib}

\begin{thebibliography}{10}\itemsep=-1pt

\bibitem{walmart}
Walmart {V}irtual {T}ry-{O}n.
\newblock \url{https://www.walmart.com/cp/virtual-try-on/4879497}.

\bibitem{bai2022single}
Shuai Bai, Huiling Zhou, Zhikang Li, Chang Zhou, and Hongxia Yang.
\newblock Single stage virtual try-on via deformable attention flows.
\newblock In {\em European Conference on Computer Vision}, pages 409--425.
  Springer, 2022.

\bibitem{binkowski2018demystifying}
Miko{\l}aj Bi{\'n}kowski, Danica~J Sutherland, Michael Arbel, and Arthur
  Gretton.
\newblock Demystifying mmd gans.
\newblock {\em arXiv preprint arXiv:1801.01401}, 2018.

\bibitem{jax2018github}
James Bradbury, Roy Frostig, Peter Hawkins, Matthew~James Johnson, Chris Leary,
  Dougal Maclaurin, George Necula, Adam Paszke, Jake Vander{P}las, Skye
  Wanderman-{M}ilne, and Qiao Zhang.
\newblock {JAX}: composable transformations of {P}ython+{N}um{P}y programs,
  2018.

\bibitem{brock2018large}
Andrew Brock, Jeff Donahue, and Karen Simonyan.
\newblock Large scale gan training for high fidelity natural image synthesis.
\newblock {\em arXiv preprint arXiv:1809.11096}, 2018.

\bibitem{choi2021viton}
Seunghwan Choi, Sunghyun Park, Minsoo Lee, and Jaegul Choo.
\newblock Viton-hd: High-resolution virtual try-on via misalignment-aware
  normalization.
\newblock In {\em Proceedings of the IEEE/CVF Conference on Computer Vision and
  Pattern Recognition}, pages 14131--14140, 2021.

\bibitem{dong2022dressing}
Xin Dong, Fuwei Zhao, Zhenyu Xie, Xijin Zhang, Daniel~K Du, Min Zheng, Xiang
  Long, Xiaodan Liang, and Jianchao Yang.
\newblock Dressing in the wild by watching dance videos.
\newblock In {\em Proceedings of the IEEE/CVF Conference on Computer Vision and
  Pattern Recognition}, pages 3480--3489, 2022.

\bibitem{dumoulin2018feature}
Vincent Dumoulin, Ethan Perez, Nathan Schucher, Florian Strub, Harm~de Vries,
  Aaron Courville, and Yoshua Bengio.
\newblock Feature-wise transformations.
\newblock {\em Distill}, 3(7):e11, 2018.

\bibitem{elfwing2018sigmoid}
Stefan Elfwing, Eiji Uchibe, and Kenji Doya.
\newblock Sigmoid-weighted linear units for neural network function
  approximation in reinforcement learning.
\newblock {\em Neural Networks}, 107:3--11, 2018.

\bibitem{ge2021parser}
Yuying Ge, Yibing Song, Ruimao Zhang, Chongjian Ge, Wei Liu, and Ping Luo.
\newblock Parser-free virtual try-on via distilling appearance flows.
\newblock In {\em Proceedings of the IEEE/CVF conference on computer vision and
  pattern recognition}, pages 8485--8493, 2021.

\bibitem{gong2019graphonomy}
Ke Gong, Yiming Gao, Xiaodan Liang, Xiaohui Shen, Meng Wang, and Liang Lin.
\newblock Graphonomy: Universal human parsing via graph transfer learning.
\newblock In {\em Proceedings of the IEEE/CVF Conference on Computer Vision and
  Pattern Recognition}, pages 7450--7459, 2019.

\bibitem{goodfellow2020generative}
Ian Goodfellow, Jean Pouget-Abadie, Mehdi Mirza, Bing Xu, David Warde-Farley,
  Sherjil Ozair, Aaron Courville, and Yoshua Bengio.
\newblock Generative adversarial networks.
\newblock {\em Communications of the ACM}, 63(11):139--144, 2020.

\bibitem{han2019clothflow}
Xintong Han, Xiaojun Hu, Weilin Huang, and Matthew~R Scott.
\newblock Clothflow: A flow-based model for clothed person generation.
\newblock In {\em Proceedings of the IEEE/CVF international conference on
  computer vision}, pages 10471--10480, 2019.

\bibitem{han2018viton}
Xintong Han, Zuxuan Wu, Zhe Wu, Ruichi Yu, and Larry~S Davis.
\newblock Viton: An image-based virtual try-on network.
\newblock In {\em Proceedings of the IEEE conference on computer vision and
  pattern recognition}, pages 7543--7552, 2018.

\bibitem{He_2022_CVPR}
Sen He, Yi-Zhe Song, and Tao Xiang.
\newblock Style-based global appearance flow for virtual try-on.
\newblock In {\em Proceedings of the IEEE/CVF Conference on Computer Vision and
  Pattern Recognition (CVPR)}, pages 3470--3479, June 2022.

\bibitem{heusel2017gans}
Martin Heusel, Hubert Ramsauer, Thomas Unterthiner, Bernhard Nessler, and Sepp
  Hochreiter.
\newblock Gans trained by a two time-scale update rule converge to a local nash
  equilibrium.
\newblock {\em Advances in neural information processing systems}, 30, 2017.

\bibitem{ho2020denoising}
Jonathan Ho, Ajay Jain, and Pieter Abbeel.
\newblock Denoising diffusion probabilistic models.
\newblock {\em Advances in Neural Information Processing Systems},
  33:6840--6851, 2020.

\bibitem{ho2022cascaded}
Jonathan Ho, Chitwan Saharia, William Chan, David~J Fleet, Mohammad Norouzi,
  and Tim Salimans.
\newblock Cascaded diffusion models for high fidelity image generation.
\newblock {\em J. Mach. Learn. Res.}, 23:47--1, 2022.

\bibitem{ho2022classifier}
Jonathan Ho and Tim Salimans.
\newblock Classifier-free diffusion guidance.
\newblock {\em arXiv preprint arXiv:2207.12598}, 2022.

\bibitem{issenhuth2020not}
Thibaut Issenhuth, J{\'e}r{\'e}mie Mary, and Cl{\'e}ment Calauz{\`e}nes.
\newblock Do not mask what you do not need to mask: a parser-free virtual
  try-on.
\newblock In {\em European Conference on Computer Vision}, pages 619--635.
  Springer, 2020.

\bibitem{johnson2016perceptual}
Justin Johnson, Alexandre Alahi, and Li Fei-Fei.
\newblock Perceptual losses for real-time style transfer and super-resolution.
\newblock In {\em European conference on computer vision}, pages 694--711.
  Springer, 2016.

\bibitem{karras2020training}
Tero Karras, Miika Aittala, Janne Hellsten, Samuli Laine, Jaakko Lehtinen, and
  Timo Aila.
\newblock Training generative adversarial networks with limited data.
\newblock {\em Advances in Neural Information Processing Systems},
  33:12104--12114, 2020.

\bibitem{karras2020analyzing}
Tero Karras, Samuli Laine, Miika Aittala, Janne Hellsten, Jaakko Lehtinen, and
  Timo Aila.
\newblock Analyzing and improving the image quality of stylegan.
\newblock In {\em Proceedings of the IEEE/CVF conference on computer vision and
  pattern recognition}, pages 8110--8119, 2020.

\bibitem{kingma2014adam}
Diederik~P Kingma and Jimmy Ba.
\newblock Adam: A method for stochastic optimization.
\newblock {\em arXiv preprint arXiv:1412.6980}, 2014.

\bibitem{lee2022hrviton}
Sangyun Lee, Gyojung Gu, Sunghyun Park, Seunghwan Choi, and Jaegul Choo.
\newblock High-resolution virtual try-on with misalignment and
  occlusion-handled conditions.
\newblock In {\em Proceedings of the European conference on computer vision
  (ECCV)}, 2022.

\bibitem{lewis2021tryongan}
Kathleen~M Lewis, Srivatsan Varadharajan, and Ira Kemelmacher-Shlizerman.
\newblock Tryongan: Body-aware try-on via layered interpolation.
\newblock {\em ACM Transactions on Graphics (TOG)}, 40(4):1--10, 2021.

\bibitem{men2020controllable}
Yifang Men, Yiming Mao, Yuning Jiang, Wei-Ying Ma, and Zhouhui Lian.
\newblock Controllable person image synthesis with attribute-decomposed gan.
\newblock In {\em Proceedings of the IEEE/CVF conference on computer vision and
  pattern recognition}, pages 5084--5093, 2020.

\bibitem{papandreou2017towards}
George Papandreou, Tyler Zhu, Nori Kanazawa, Alexander Toshev, Jonathan
  Tompson, Chris Bregler, and Kevin Murphy.
\newblock Towards accurate multi-person pose estimation in the wild.
\newblock In {\em Proceedings of the IEEE conference on computer vision and
  pattern recognition}, pages 4903--4911, 2017.

\bibitem{radford2021learning}
Alec Radford, Jong~Wook Kim, Chris Hallacy, Aditya Ramesh, Gabriel Goh,
  Sandhini Agarwal, Girish Sastry, Amanda Askell, Pamela Mishkin, Jack Clark,
  et~al.
\newblock Learning transferable visual models from natural language
  supervision.
\newblock In {\em International Conference on Machine Learning}, pages
  8748--8763. PMLR, 2021.

\bibitem{ramesh2022hierarchical}
Aditya Ramesh, Prafulla Dhariwal, Alex Nichol, Casey Chu, and Mark Chen.
\newblock Hierarchical text-conditional image generation with clip latents.
\newblock {\em arXiv preprint arXiv:2204.06125}, 2022.

\bibitem{reda2022film}
Fitsum Reda, Janne Kontkanen, Eric Tabellion, Deqing Sun, Caroline Pantofaru,
  and Brian Curless.
\newblock Film: Frame interpolation for large motion.
\newblock In {\em The European Conference on Computer Vision (ECCV)}, 2022.

\bibitem{ren2022neural}
Yurui Ren, Xiaoqing Fan, Ge Li, Shan Liu, and Thomas~H Li.
\newblock Neural texture extraction and distribution for controllable person
  image synthesis.
\newblock In {\em Proceedings of the IEEE/CVF Conference on Computer Vision and
  Pattern Recognition}, pages 13535--13544, 2022.

\bibitem{rombach2022high}
Robin Rombach, Andreas Blattmann, Dominik Lorenz, Patrick Esser, and Bj{\"o}rn
  Ommer.
\newblock High-resolution image synthesis with latent diffusion models.
\newblock In {\em Proceedings of the IEEE/CVF Conference on Computer Vision and
  Pattern Recognition}, pages 10684--10695, 2022.

\bibitem{ronneberger2015u}
Olaf Ronneberger, Philipp Fischer, and Thomas Brox.
\newblock U-net: Convolutional networks for biomedical image segmentation.
\newblock In {\em International Conference on Medical image computing and
  computer-assisted intervention}, pages 234--241. Springer, 2015.

\bibitem{ruiz2022dreambooth}
Nataniel Ruiz, Yuanzhen Li, Varun Jampani, Yael Pritch, Michael Rubinstein, and
  Kfir Aberman.
\newblock Dreambooth: Fine tuning text-to-image diffusion models for
  subject-driven generation.
\newblock {\em arXiv preprint arXiv:2208.12242}, 2022.

\bibitem{saharia2022palette}
Chitwan Saharia, William Chan, Huiwen Chang, Chris Lee, Jonathan Ho, Tim
  Salimans, David Fleet, and Mohammad Norouzi.
\newblock Palette: Image-to-image diffusion models.
\newblock In {\em ACM SIGGRAPH 2022 Conference Proceedings}, pages 1--10, 2022.

\bibitem{saharia2022photorealistic}
Chitwan Saharia, William Chan, Saurabh Saxena, Lala Li, Jay Whang, Emily
  Denton, Seyed Kamyar~Seyed Ghasemipour, Burcu~Karagol Ayan, S~Sara Mahdavi,
  Rapha~Gontijo Lopes, et~al.
\newblock Photorealistic text-to-image diffusion models with deep language
  understanding.
\newblock {\em Advances in Neural Information Processing Systems}, 2022.

\bibitem{saharia2022image}
Chitwan Saharia, Jonathan Ho, William Chan, Tim Salimans, David~J Fleet, and
  Mohammad Norouzi.
\newblock Image super-resolution via iterative refinement.
\newblock {\em IEEE Transactions on Pattern Analysis and Machine Intelligence},
  2022.

\bibitem{sohl2015deep}
Jascha Sohl-Dickstein, Eric Weiss, Niru Maheswaranathan, and Surya Ganguli.
\newblock Deep unsupervised learning using nonequilibrium thermodynamics.
\newblock In {\em International Conference on Machine Learning}, pages
  2256--2265. PMLR, 2015.

\bibitem{song2020denoising}
Jiaming Song, Chenlin Meng, and Stefano Ermon.
\newblock Denoising diffusion implicit models.
\newblock {\em arXiv preprint arXiv:2010.02502}, 2020.

\bibitem{song2019generative}
Yang Song and Stefano Ermon.
\newblock Generative modeling by estimating gradients of the data distribution.
\newblock {\em Advances in Neural Information Processing Systems}, 32, 2019.

\bibitem{vaswani2017attention}
Ashish Vaswani, Noam Shazeer, Niki Parmar, Jakob Uszkoreit, Llion Jones,
  Aidan~N Gomez, {\L}ukasz Kaiser, and Illia Polosukhin.
\newblock Attention is all you need.
\newblock {\em Advances in neural information processing systems}, 30, 2017.

\bibitem{wang2018toward}
Bochao Wang, Huabin Zheng, Xiaodan Liang, Yimin Chen, Liang Lin, and Meng Yang.
\newblock Toward characteristic-preserving image-based virtual try-on network.
\newblock In {\em Proceedings of the European conference on computer vision
  (ECCV)}, pages 589--604, 2018.

\bibitem{watson2022novel}
Daniel Watson, William Chan, Ricardo Martin-Brualla, Jonathan Ho, Andrea
  Tagliasacchi, and Mohammad Norouzi.
\newblock Novel view synthesis with diffusion models.
\newblock {\em arXiv preprint arXiv:2210.04628}, 2022.

\bibitem{wu2018group}
Yuxin Wu and Kaiming He.
\newblock Group normalization.
\newblock In {\em Proceedings of the European conference on computer vision
  (ECCV)}, pages 3--19, 2018.

\bibitem{Yang_2022_CVPR}
Han Yang, Xinrui Yu, and Ziwei Liu.
\newblock Full-range virtual try-on with recurrent tri-level transform.
\newblock In {\em Proceedings of the IEEE/CVF Conference on Computer Vision and
  Pattern Recognition (CVPR)}, pages 3460--3469, June 2022.

\bibitem{yang2020towards}
Han Yang, Ruimao Zhang, Xiaobao Guo, Wei Liu, Wangmeng Zuo, and Ping Luo.
\newblock Towards photo-realistic virtual try-on by adaptively
  generating-preserving image content.
\newblock In {\em Proceedings of the IEEE/CVF conference on computer vision and
  pattern recognition}, pages 7850--7859, 2020.

\bibitem{yu2019vtnfp}
Ruiyun Yu, Xiaoqi Wang, and Xiaohui Xie.
\newblock Vtnfp: An image-based virtual try-on network with body and clothing
  feature preservation.
\newblock In {\em Proceedings of the IEEE/CVF international conference on
  computer vision}, pages 10511--10520, 2019.

\bibitem{zhang2021pise}
Jinsong Zhang, Kun Li, Yu-Kun Lai, and Jingyu Yang.
\newblock Pise: Person image synthesis and editing with decoupled gan.
\newblock In {\em Proceedings of the IEEE/CVF Conference on Computer Vision and
  Pattern Recognition}, pages 7982--7990, 2021.

\bibitem{zhu2020deformable}
Xizhou Zhu, Weijie Su, Lewei Lu, Bin Li, Xiaogang Wang, and Jifeng Dai.
\newblock Deformable detr: Deformable transformers for end-to-end object
  detection.
\newblock {\em arXiv preprint arXiv:2010.04159}, 2020.

\end{thebibliography}
}

\newpage
\clearpage
\appendix

\makeatletter

\section*{Appendix}

\begin{figure*}[h]
\begin{center}
  \includegraphics[width=1.0\linewidth]{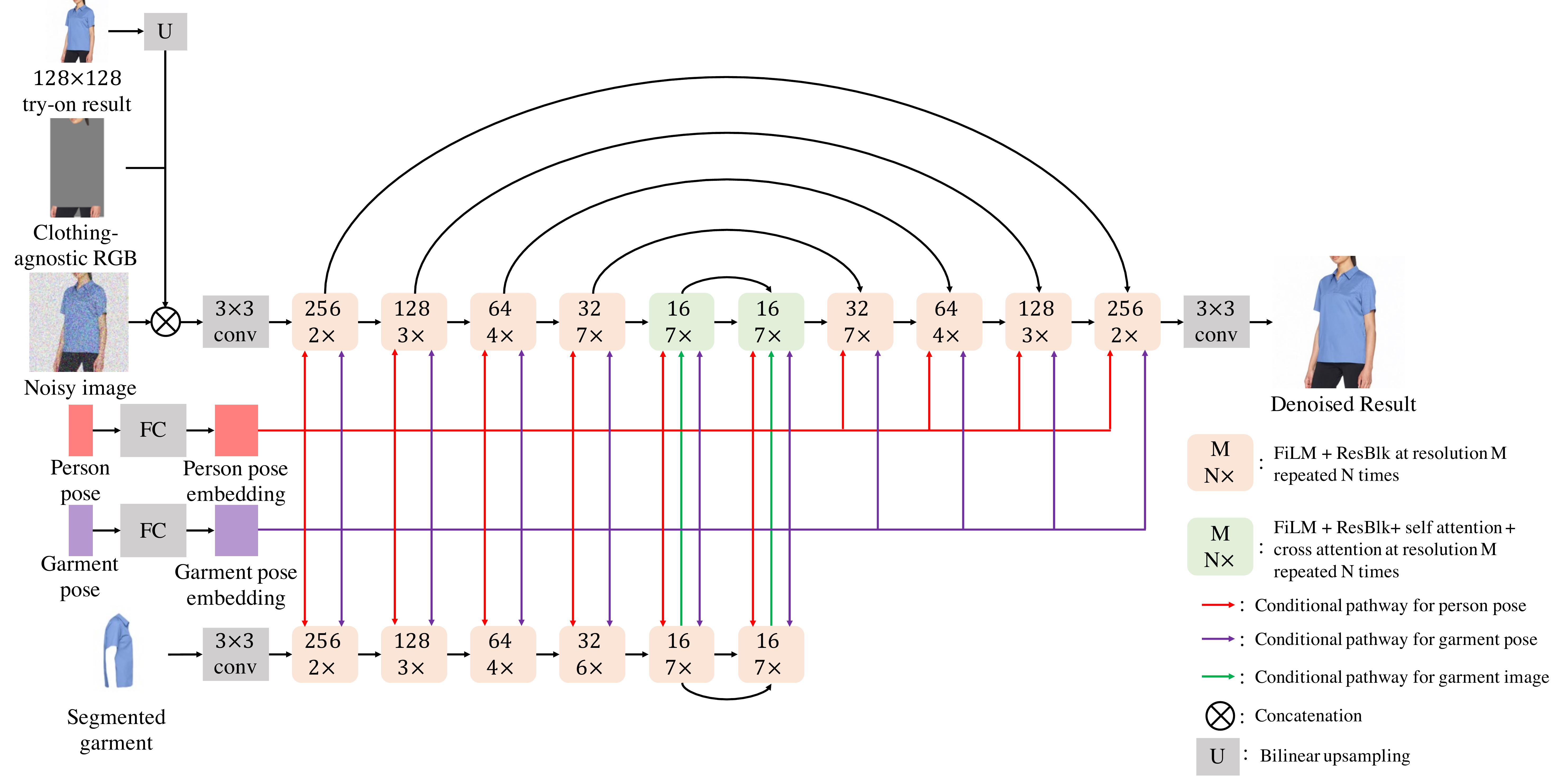}
\end{center}
  \caption{Architecture of $256\mathord\times\mathord256$ \unet{}.}
\label{fig:suppl_parralel_unet_256}
\end{figure*}

\section{Implementation Details}
\subsection{\unet{}}
Fig.~\ref{fig:suppl_parralel_unet_256} provides the architecture of $256\mathord\times\mathord256$ \unet{}. Compared to the $128\mathord\times\mathord128$ version, $256\mathord\times\mathord256$ \unet{} makes the following changes: 1) In addition to the try-on conditional inputs  $\mathbf{c}_\text{tryon}$, the $256\mathord\times\mathord256$ \unet takes as input the try-on result $I_{\text{tr}}^{128}$, which is first bilinearly upsampled to $256\mathord\times\mathord256$, and then concatenated to the noisy image $\mathbf{z}_t$; 2) the self attention and cross attention modules only happen at $16\mathord\times\mathord16$ resolution; 3) extra UNet blocks at $256\mathord\times\mathord256$ resolution are used; 4) the repeated times of UNet blocks are different as indicated by the Figures.

For both $128\mathord\times\mathord128$ and $256\mathord\times\mathord256$ \unet{}, normalization layers are parametrized as Group Normalization~\cite{wu2018group}. The number of group is set to $\min(32, \lfloor \frac{C}{4} \rfloor)$, where $C$ is the number of channels for input features. The non-linear activation is set to swish~\cite{elfwing2018sigmoid} across the whole model. The residual blocks used in each scale have a main pathway of $\text{GroupNorm}\mathord\rightarrow\mathord\text{swish}\mathord\rightarrow\mathord\text{conv}\mathord\rightarrow\mathord\text{GroupNorm}\mathord\rightarrow\mathord\text{swish}\mathord\rightarrow\mathord\text{conv}$. The input to the residual block is processed by a separate convolution layer and added to added to the output of the main pathway as the skip connection. The number of feature channels for UNet blocks in $128\mathord\times\mathord128$ \unet{} is set to $128, 256, 512, 1024$ for resolution $128, 64, 32, 16$ respectively. The number of feature channels for UNet blocks in $256\mathord\times\mathord256$ \unet{} is set to $128, 128, 256, 512, 1024$ for resolution $256, 128, 64, 32, 16$ respectively. The positional encodings of diffusion timstep $t$ and noise augmentation levels $\mathbf{t}_{\text{na}}$ are not shown in the figures for cleaner visualization. They are used for FiLM~\cite{dumoulin2018feature} as described in Sec.~\ref{subsec::parallel_unet}. The $128\mathord\times\mathord128$ \unet{} has 1.13B parameters in total while the $256\mathord\times\mathord256$ \unet{} has 1.06B parameters.

\subsection{Training and Inference}
TryOnDiffusion was implemented in JAX~\cite{jax2018github}. All three diffusion models are trained on 32 TPU-v4 chips for 500K iterations (around 3 days for each diffusion model). After trained, we run the inference of the whole pipeline on 4 TPU-v4 chips with batch size 4, which takes around 18 seconds for one batch.

\section{Additional Results}

\begin{figure*}
\begin{center}
  \includegraphics[width=1.0\linewidth]{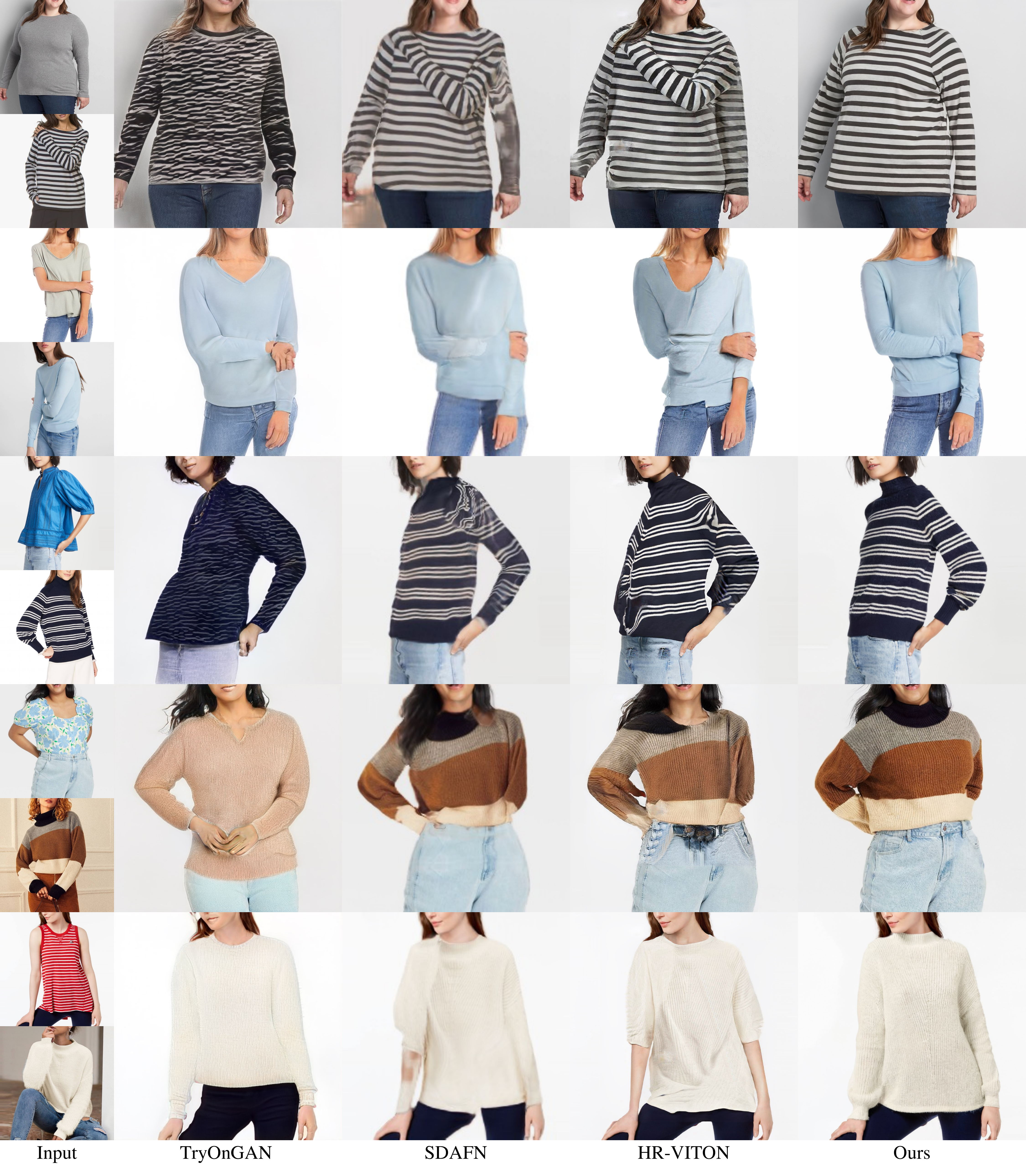}
\end{center}
\vspace{-5mm}
  \caption{Comparison with TryOnGAN~\cite{lewis2021tryongan}, SDAFN~\cite{bai2022single} and HR-VITON~\cite{lee2022hrviton} on challenging cases for women. Compared to baselines, TryOnDiffusion can preserve garment details for heavy occlusions as well as extreme body pose and shape differences. Please zoom in to see details.}
\label{fig:suppl_woman_hard_pose}
\end{figure*}

\begin{figure*}
\begin{center}
  \includegraphics[width=1.0\linewidth]{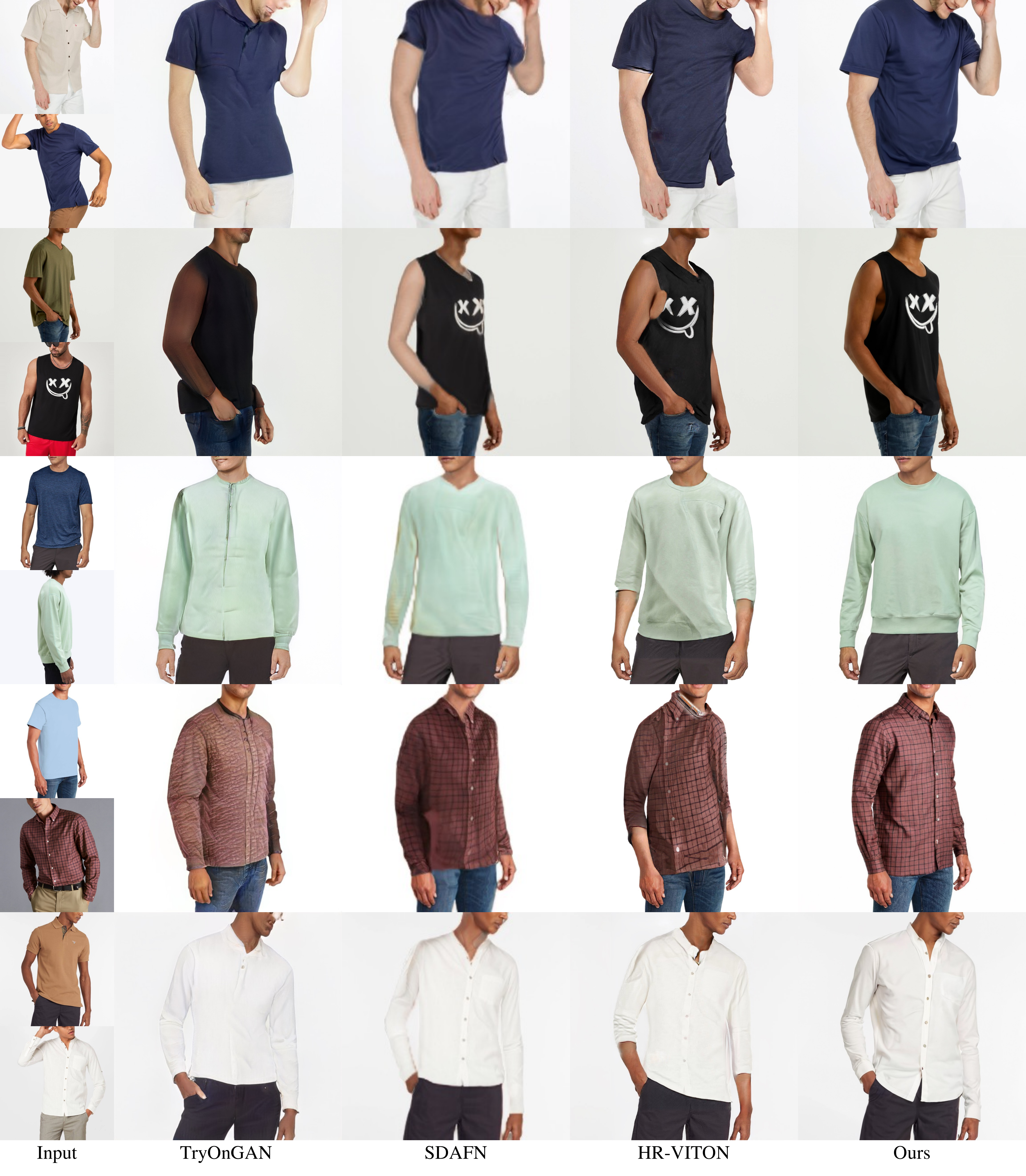}
\end{center}
\vspace{-5mm}
  \caption{Comparison with TryOnGAN~\cite{lewis2021tryongan}, SDAFN~\cite{bai2022single} and HR-VITON~\cite{lee2022hrviton} on challenging cases for men. Compared to baselines, TryOnDiffusion can preserve garment details for heavy occlusions as well as extreme body pose and shape differences. Please zoom in to see details.}
\label{fig:suppl_man_hard_pose}
\end{figure*}

\begin{figure*}
\begin{center}
  \includegraphics[width=1.0\linewidth]{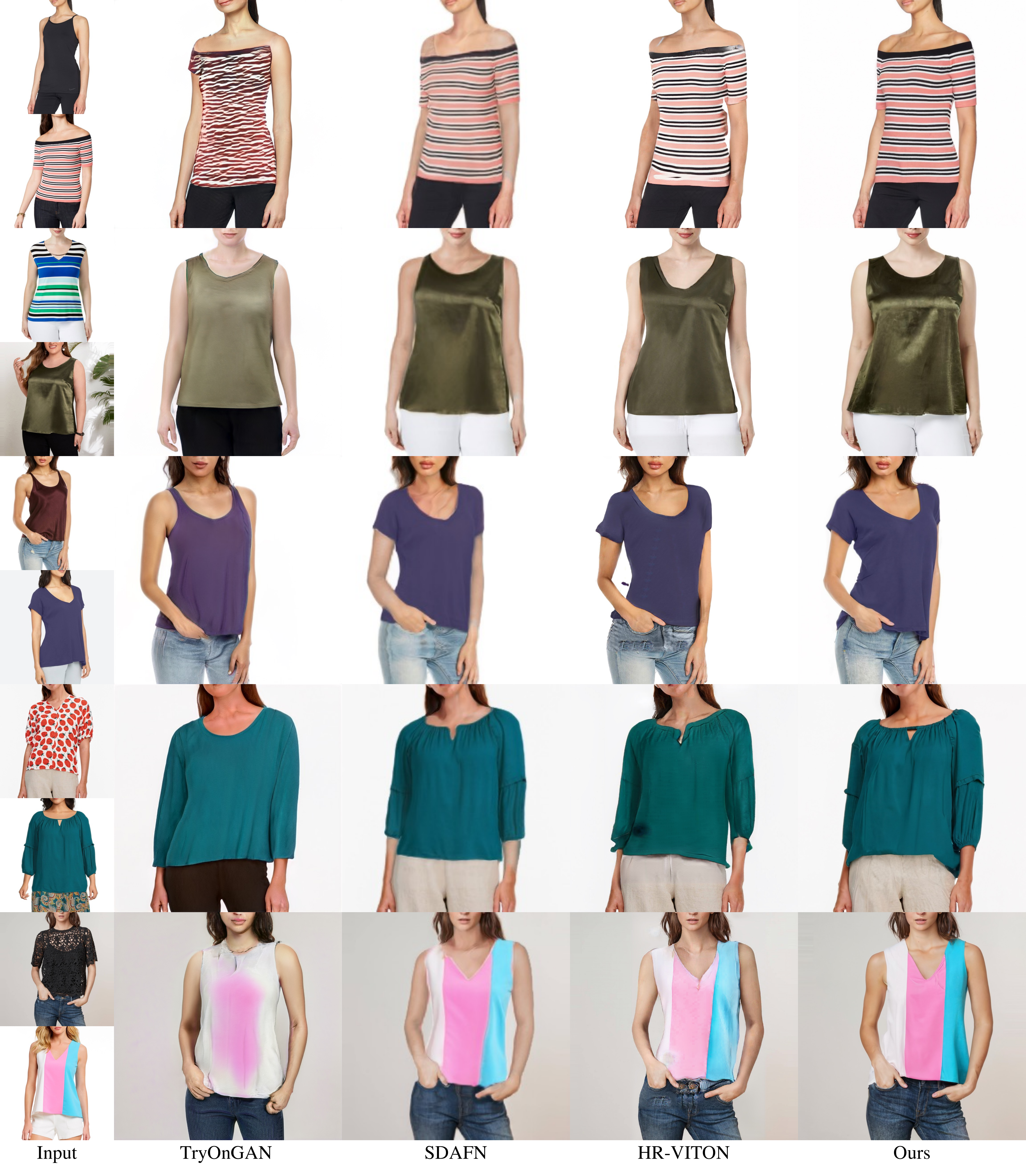}
\end{center}
\vspace{-5mm}
  \caption{Comparison with TryOnGAN~\cite{lewis2021tryongan}, SDAFN~\cite{bai2022single} and HR-VITON~\cite{lee2022hrviton} on simple cases for women. We select input pairs with minimum garment warp and simple texture pattern. Baseline methods perform better for simple cases than for challenging cases. However, our method is still better at garment detail preservation and blending (of person and garment). Please zoom in to see details.}
\label{fig:suppl_woman_easy_pose}
\end{figure*}

\begin{figure*}
\begin{center}
  \includegraphics[width=1.0\linewidth]{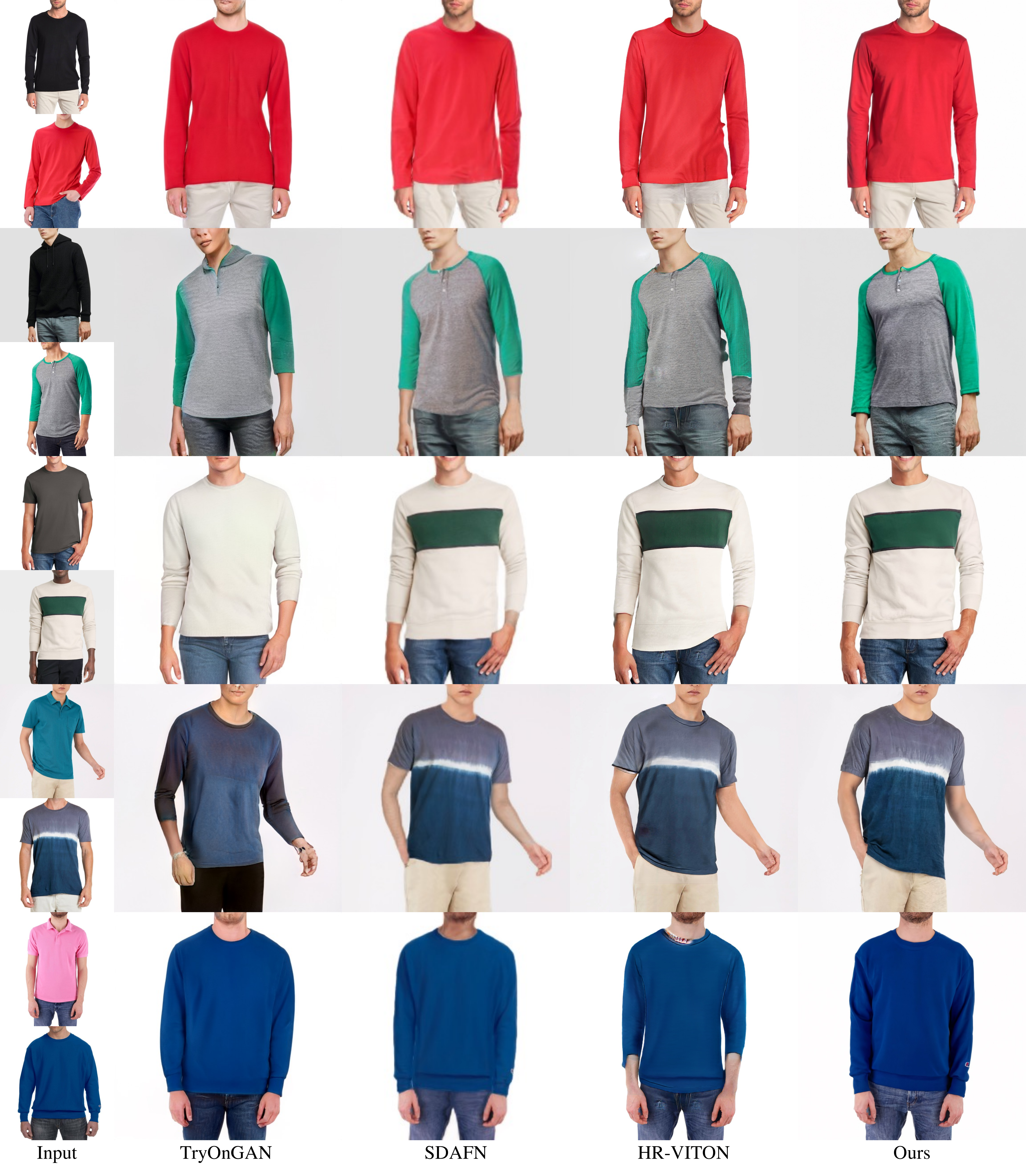}
\end{center}
\vspace{-5mm}
  \caption{Comparison with TryOnGAN~\cite{lewis2021tryongan}, SDAFN~\cite{bai2022single} and HR-VITON~\cite{lee2022hrviton} on simple cases for men. We select input pairs with minimum garment warp and simple texture pattern. Baseline methods perform better for simple cases than for challenging cases. However, our method is still better at garment detail preservation and blending (of person and garment). Please zoom in to see details.}
\label{fig:suppl_man_easy_pose}
\end{figure*}

\begin{figure*}
\begin{center}
  \includegraphics[width=1.0\linewidth]{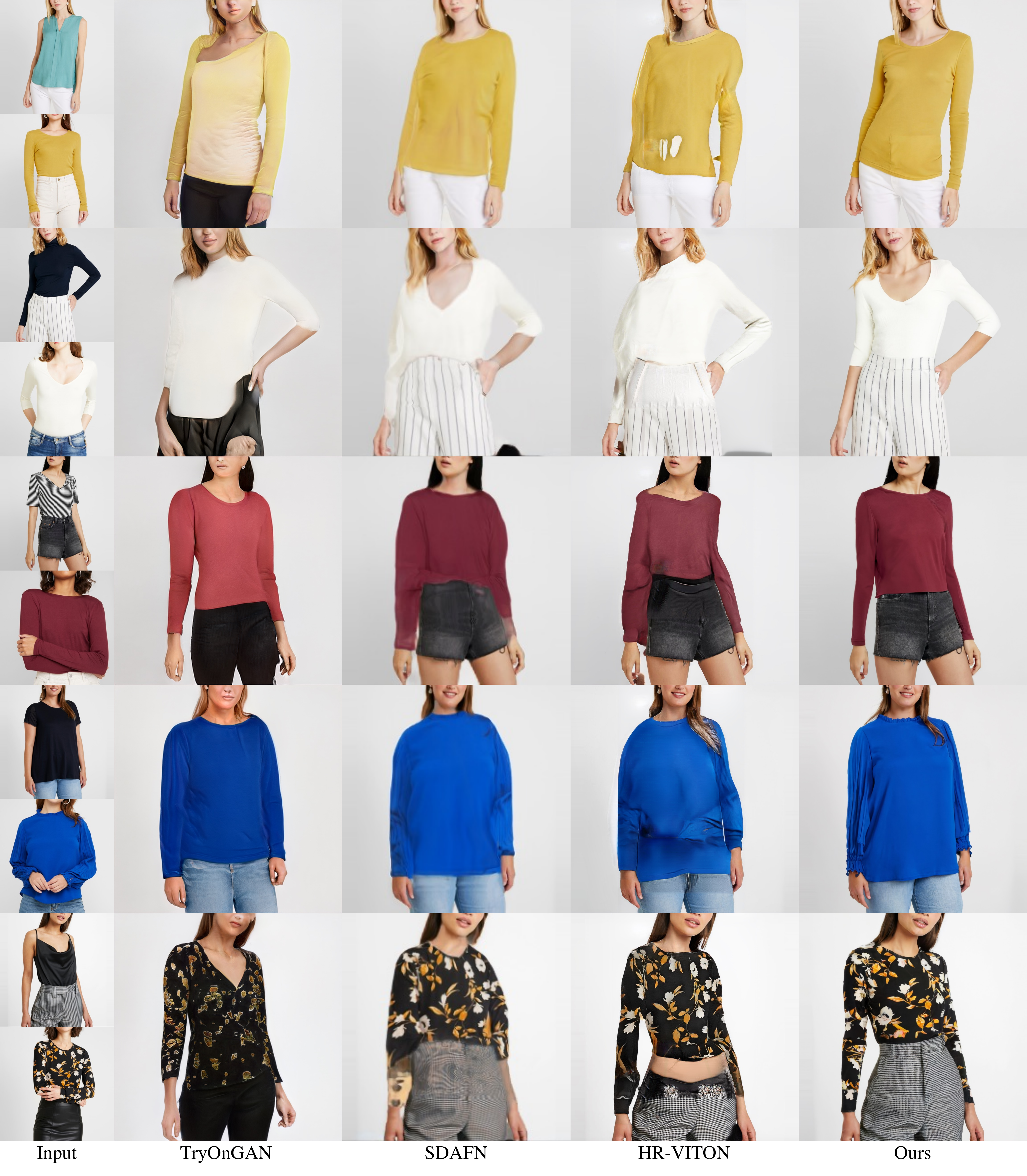}
\end{center}
\vspace{-5mm}
  \caption{Comparison with state-of-the-art methods on VITON-HD unpaired testing dataset~\cite{choi2021viton}. All methods were trained on the same 4M dataset and tested on VITON-HD. Please zoom in to see details}
\label{fig:suppl_vitonhd_dataset}
\end{figure*}

In Fig.~\ref{fig:suppl_woman_hard_pose} and \ref{fig:suppl_man_hard_pose}, we provide qualitative comparison to state-of-the-art methods on challenging cases. We select input pairs from our 6K testing dataset with heavy occlusions and extreme body pose and shape differences. We can see that our method can generate more realistic results compared to baselines.
In Fig.~\ref{fig:suppl_woman_easy_pose} and \ref{fig:suppl_man_easy_pose}, we provide qualitative comparison to state-of-the-art methods on simple cases. We select input pairs from our 6K test dataset with minimum garment warp and simple texture pattern. Baseline methods perform better for simple cases than for challenging cases. However, our method is still better at garment detail preservation and blending (of person and garment).
In Fig.~\ref{fig:suppl_vitonhd_dataset}, we provide more qualitative results on the VITON-HD unpaired testing dataset.

\begin{figure*}
\begin{center}
   \includegraphics[width=1.0\linewidth]{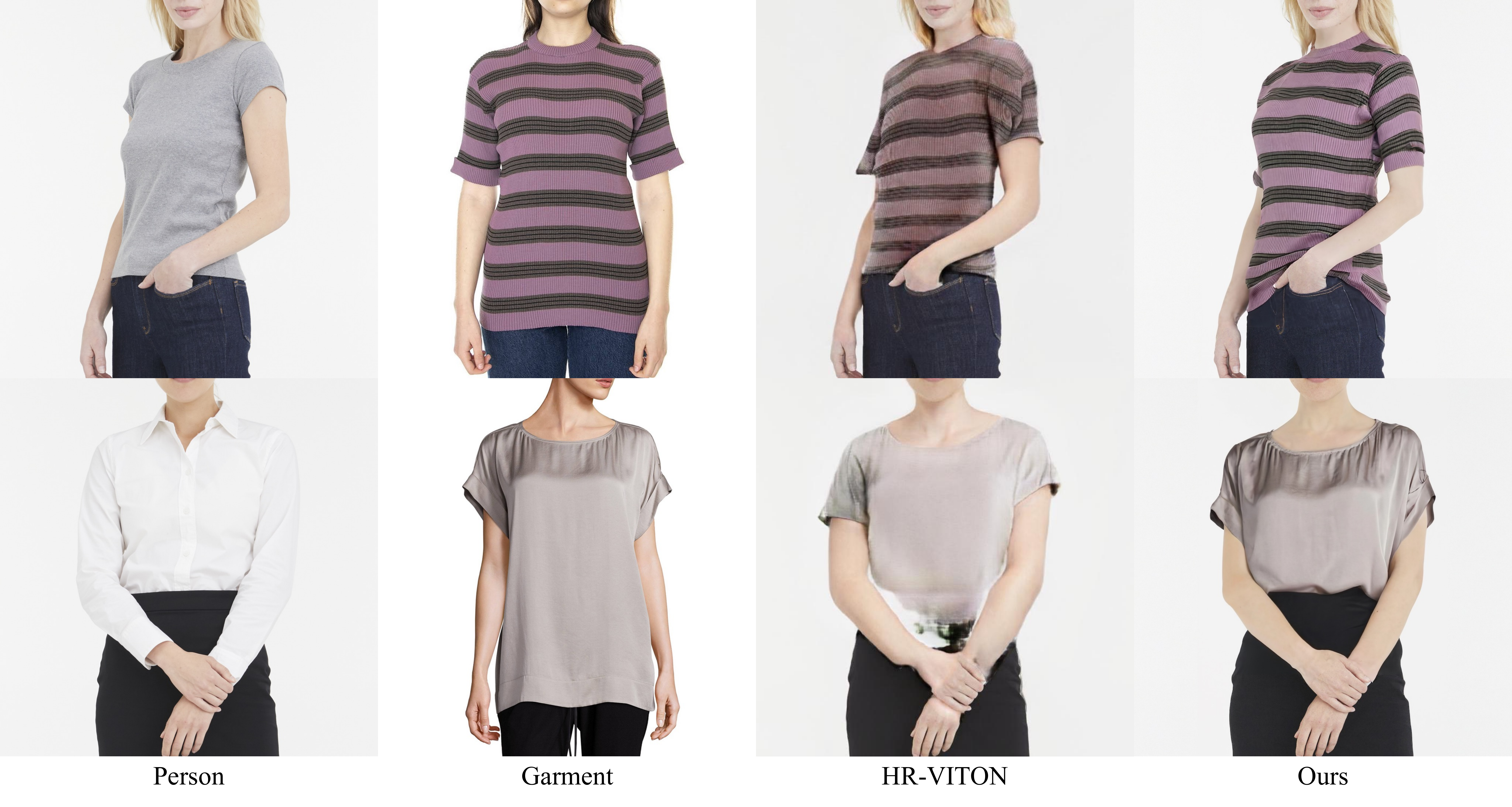}
\end{center}
\vspace{-6mm}
\caption{Comparison with HR-VITON released checkpoints for frontal garment (optimal for HR-VITON). Please zoom in to see details.
}
\label{fig:suppl_hrviton_release}
\end{figure*}

For fair comparison, we run a new user study to compare SDAFN~\cite{bai2022single} vs  our method at SDAFN's  $256\times256$ resolution. To generate a $256\times256$ image with our method, we only run inference on the first two stages of our cascaded diffusion models and ignore the $256\mathord\times\mathord256\mathord\rightarrow\mathord1024\mathord\times\mathord1024$ SR diffusion. Table~\ref{table:user_study_256} shows results consistent with the user study reported in the paper. We also compare to HR-VITON~\cite{lee2022hrviton} using their released checkpoints. Note that original HR-VTION is trained on frontal garment images, so we select input garments satisfying this constraint to avoid unfair comparison. Fig.~\ref{fig:suppl_hrviton_release} shows that our method is still better than HR-VITON under its optimal cases using its released checkpoints.

\begin{figure*}
\begin{center}
  \includegraphics[width=1.0\linewidth]{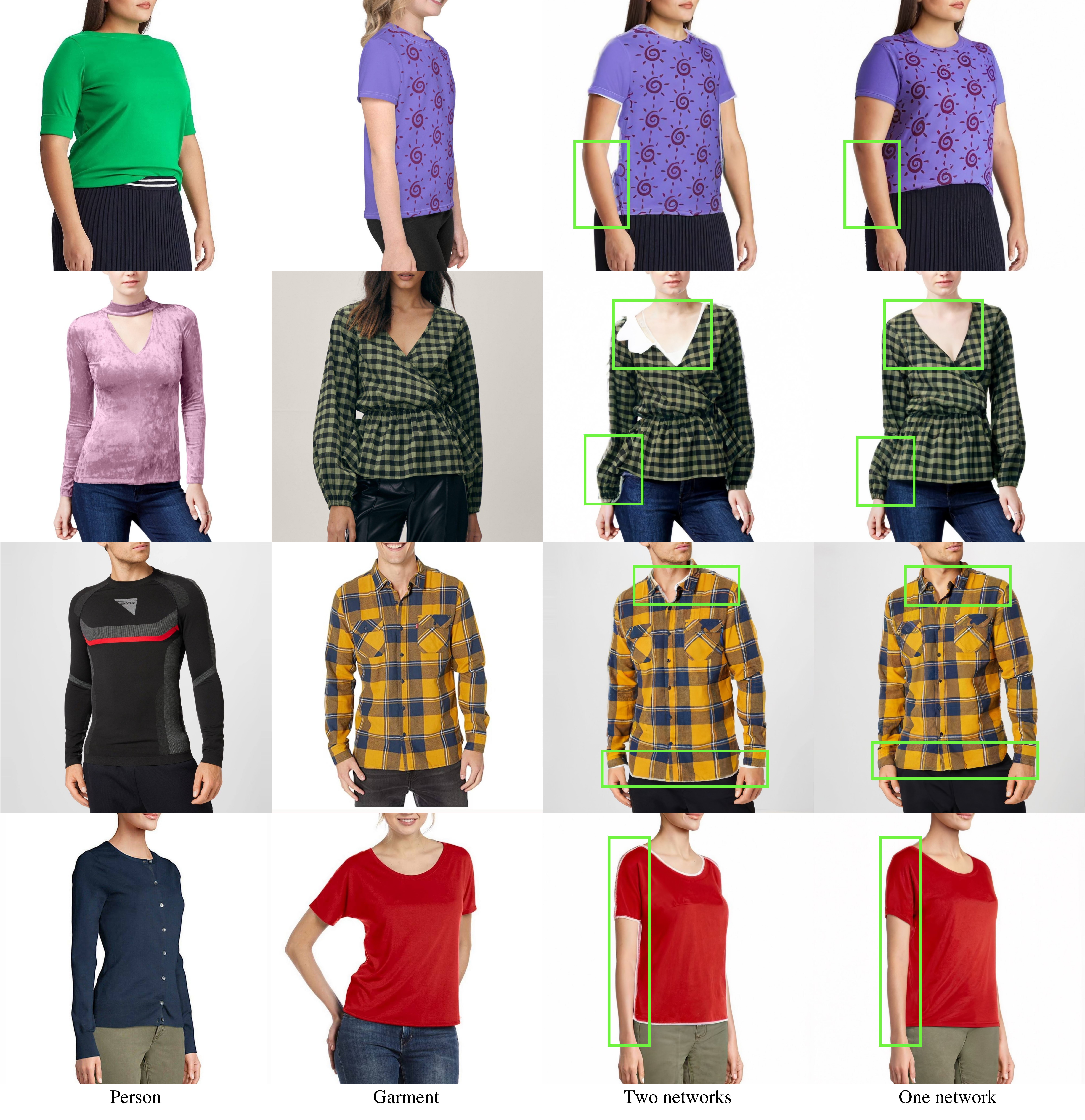}
\end{center}
  \caption{Combining warp and blend vs sequencing
two tasks. Two networks (column 3) represent sequencing two tasks. One network (column 4) represents combining warp and blend. Green boxes highlight differences, please zoom in to see details.}
\label{fig:suppl_twobd_vs_onebd}
\end{figure*}

\begin{figure*}
\begin{center}
  \includegraphics[width=1.0\linewidth]{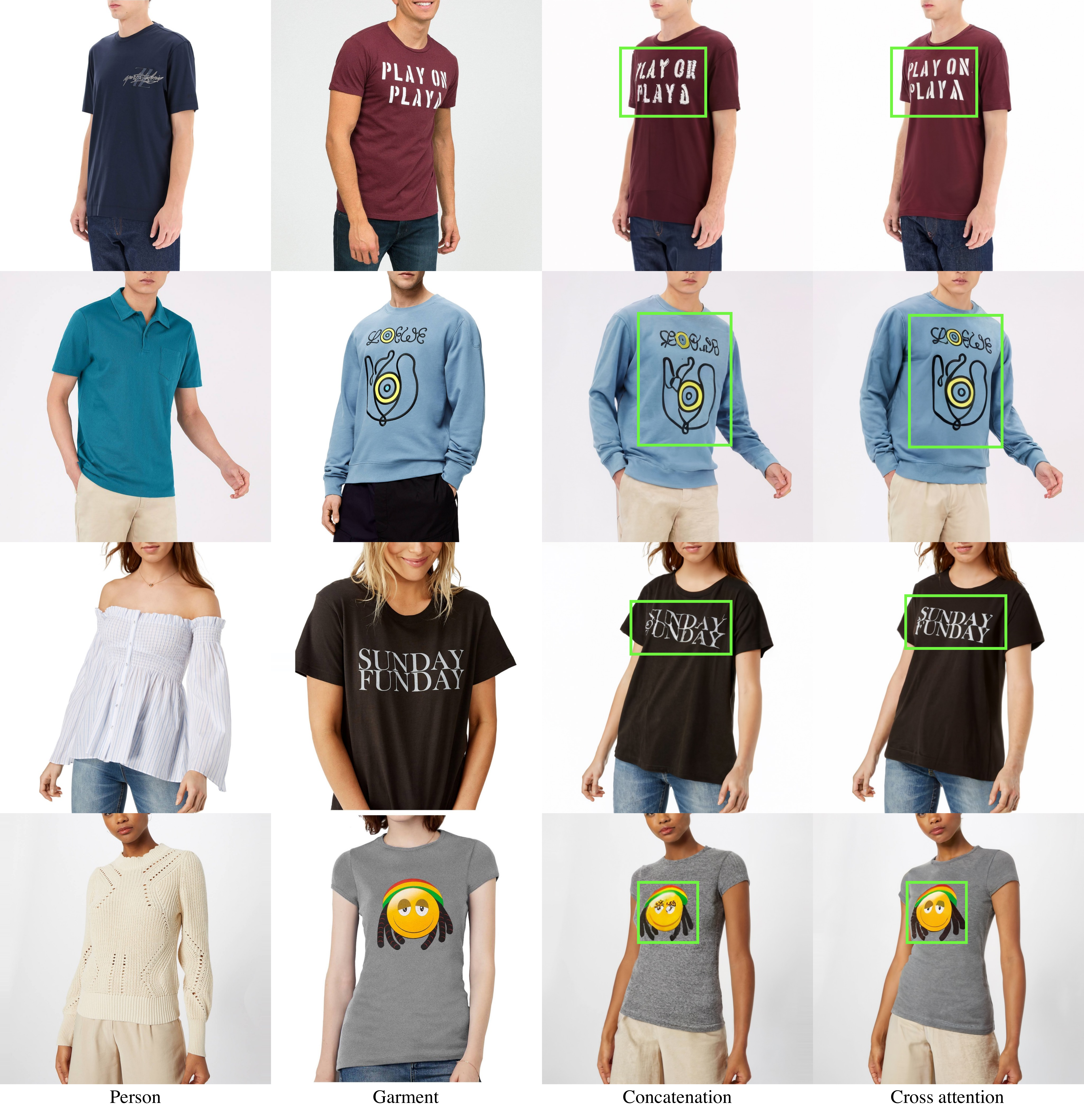}
\end{center}
  \caption{Cross attention vs concatenation for implicit
warping. Green boxes highlight differences, please zoom in to see details.}
\label{fig:suppl_concat_vs_xattn}
\end{figure*}

Table~\ref{table:quantitative_ablation} reports quantitative results for ablation studies. Fig.~\ref{fig:suppl_twobd_vs_onebd} visualizes more examples for the ablation study of combining warp and blend versus sequencing the tasks. Fig.~\ref{fig:suppl_concat_vs_xattn} provides more qualitative comparisons between concatenation and cross attention for implicit warping.

\begin{figure*}
\begin{center}
   \includegraphics[width=1.0\linewidth]{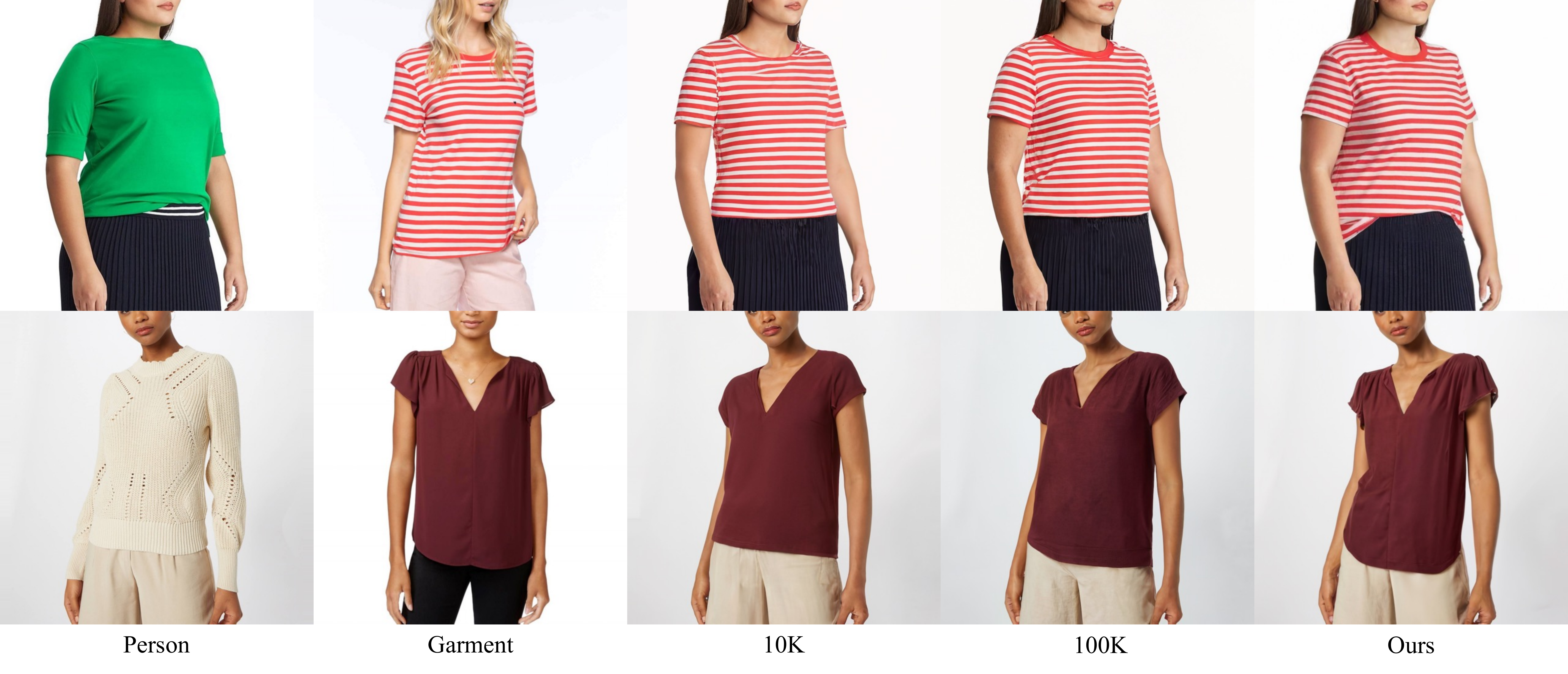}
\end{center}
\vspace{-6mm}
\caption{Quanlitative results for effects of the training set size. Please zoom in to see details.}
\label{fig:suppl_trainset_size}
\end{figure*}

We further investigate the effect of the training dataset size. We retrained our method from scratch on 10K and 100K random pairs from our 4M set and report quantitative results (FID and KID) on two different test sets in Table~\ref{table:quantitative_dataset_size}. Fig.~\ref{fig:suppl_trainset_size} also shows visual results for our models trained on different dataset sizes.

\begin{figure*}
\begin{center}
  \includegraphics[width=1.0\linewidth]{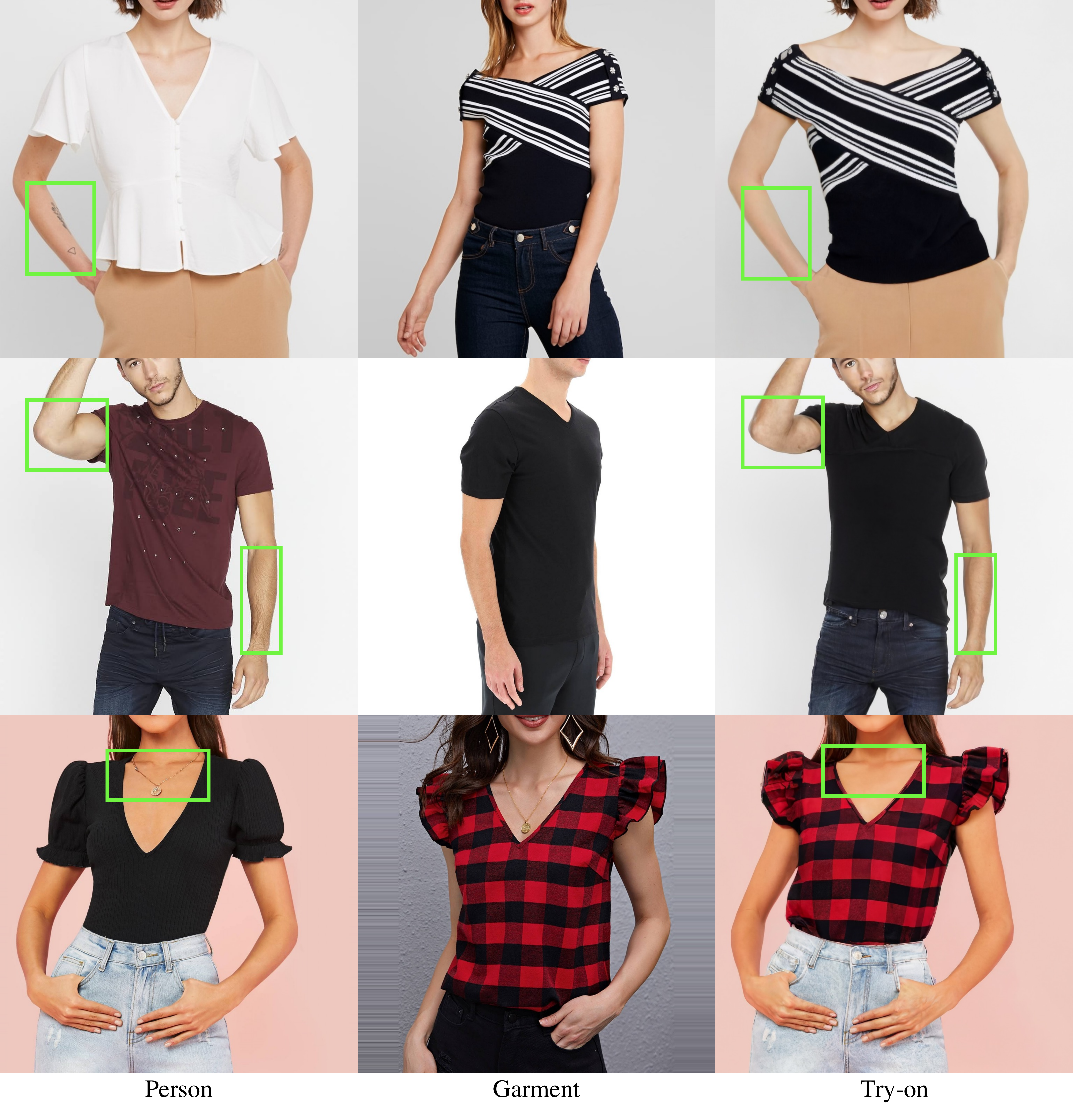}
\end{center}
  \caption{Failure cases. Clothing-agnostic RGB image removes part of the identity information from the target person, e.g., tattoos (row one), muscle structure (row two), fine hair on the skin (row two) and accessories (row three).}
\label{fig:suppl_failure_cases}
\end{figure*}

\begin{figure*}
\begin{center}
   \includegraphics[width=0.85\linewidth]{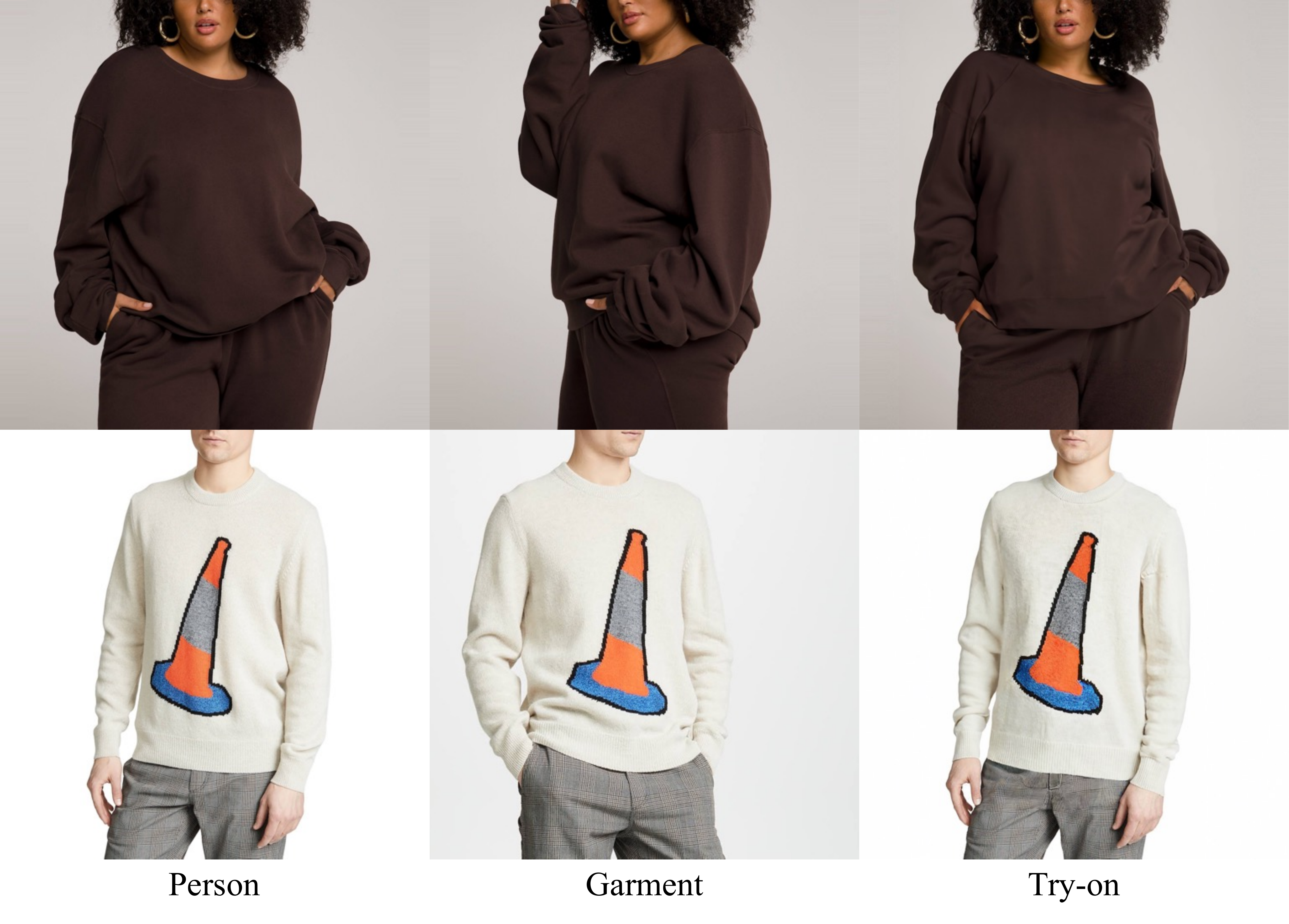}
\end{center}
\vspace{-6mm}
\caption{Qualitative results on paired unseen test samples. Please zoom in to see details.}
\label{fig:suppl_paired_testing}
\end{figure*}

In Fig.~\ref{fig:failure_case} of the main paper, we provide failure cases due to erroneous garment segmentation and garment leaks in the clothing-agnostic RGB image. In Fig.~\ref{fig:suppl_failure_cases}, we provide more failure cases of our method. The main problem lies in the clothing-agnostic RGB image. Specifically, it removes part of the identity information from the target person, e.g., tattoos (row one), muscle structure (row two), fine hair on the skin (row two) and accessories (row three). To better visualize the difference in person identity, Fig.~\ref{fig:suppl_paired_testing} provides try-on results on paired unseen test samples, where groundtruth is available.

\begin{figure*}
\begin{center}
   \includegraphics[width=0.85\linewidth]{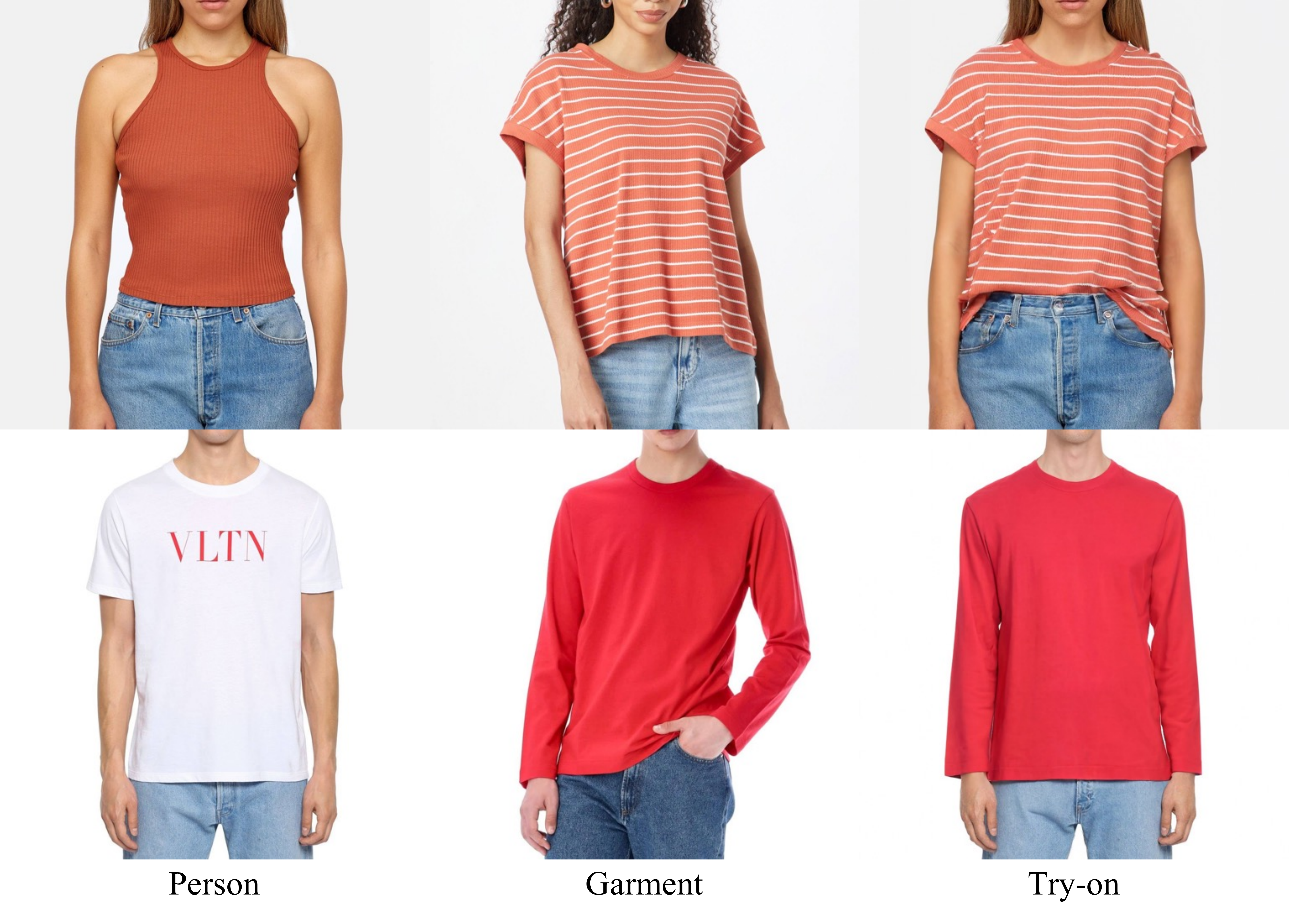}
\end{center}
\vspace{-6mm}
\caption{Try-on results for input person wearing garment with no folds, and input garment with folds.}
\label{fig:suppl_puffy}
\end{figure*}

\begin{figure*}
\begin{center}
  \includegraphics[width=1.0\linewidth]{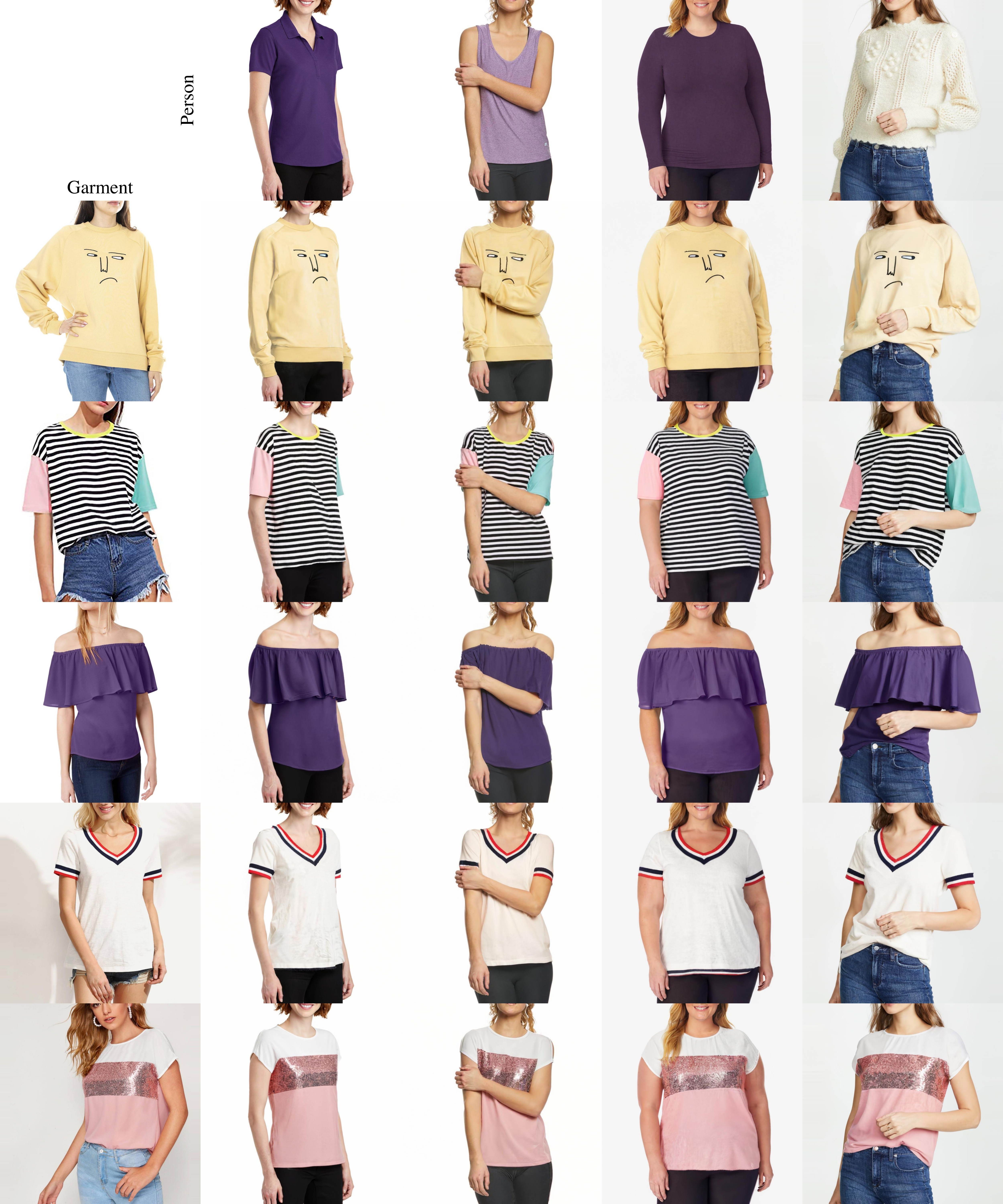}
\end{center}
  \caption{4 women trying on 5 garments.}
\label{fig:suppl_diverse_tryon_woman}
\end{figure*}

\begin{figure*}
\begin{center}
  \includegraphics[width=1.0\linewidth]{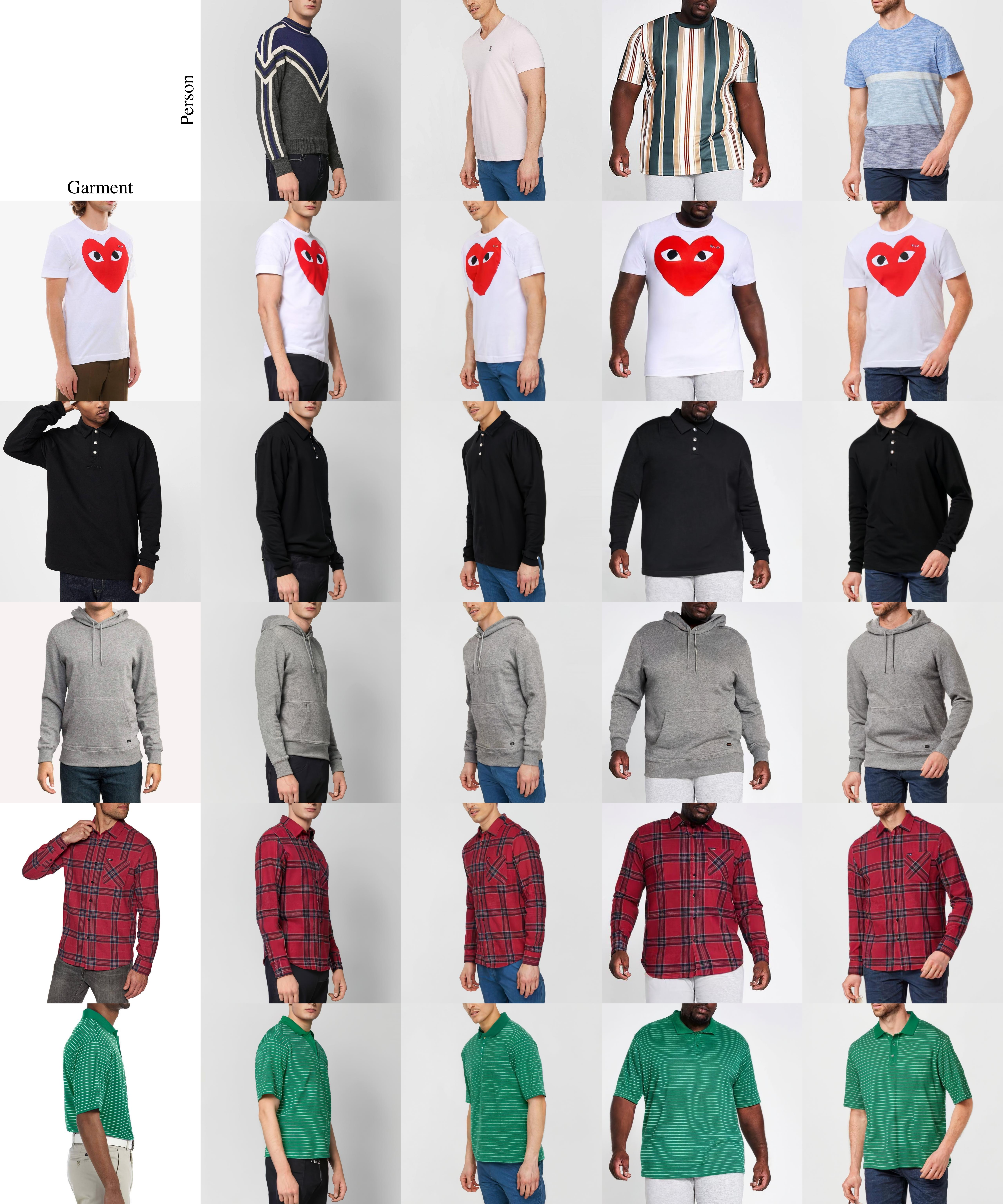}
\end{center}
  \caption{4 men trying on 5 garments.}
\label{fig:suppl_diverse_tryon_man}
\end{figure*}

Fig.~\ref{fig:suppl_puffy} shows try-on results for a challenging case, where input person wearing garment with no folds, and input garment with folds. We can see that our method can generate realistic folds according to the person pose instead of copying folds from the garment input. Fig.~\ref{fig:suppl_diverse_tryon_woman} and \ref{fig:suppl_diverse_tryon_man} show TryOnDiffusion results on variety of people and garments for both men and women.

\begin{figure*}
\begin{center}
  \includegraphics[width=1.0\linewidth]{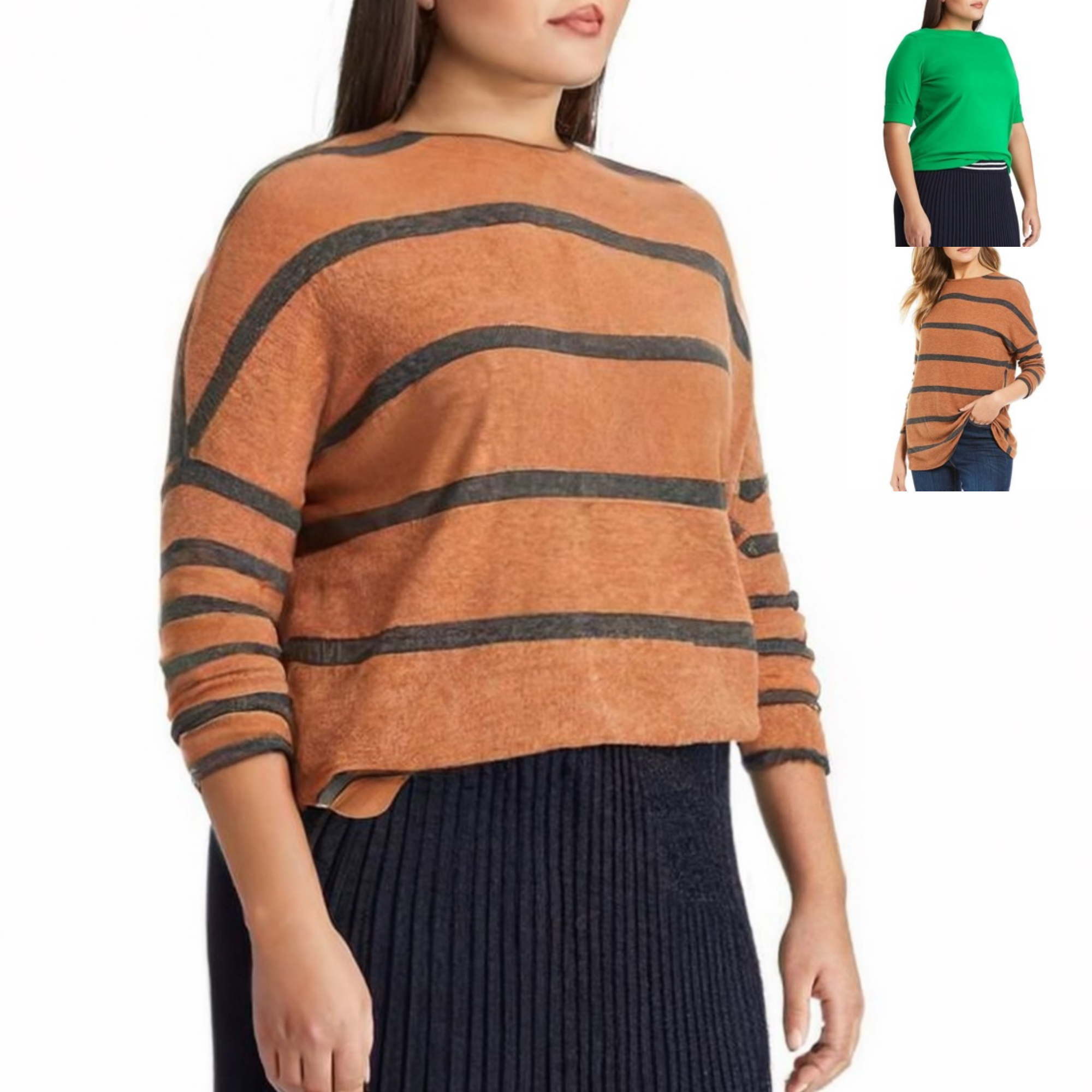}
\end{center}
  \caption{Larger version of teaser.}
\label{fig:suppl_teaser_woman1}
\end{figure*}

\begin{figure*}
\begin{center}
  \includegraphics[width=1.0\linewidth]{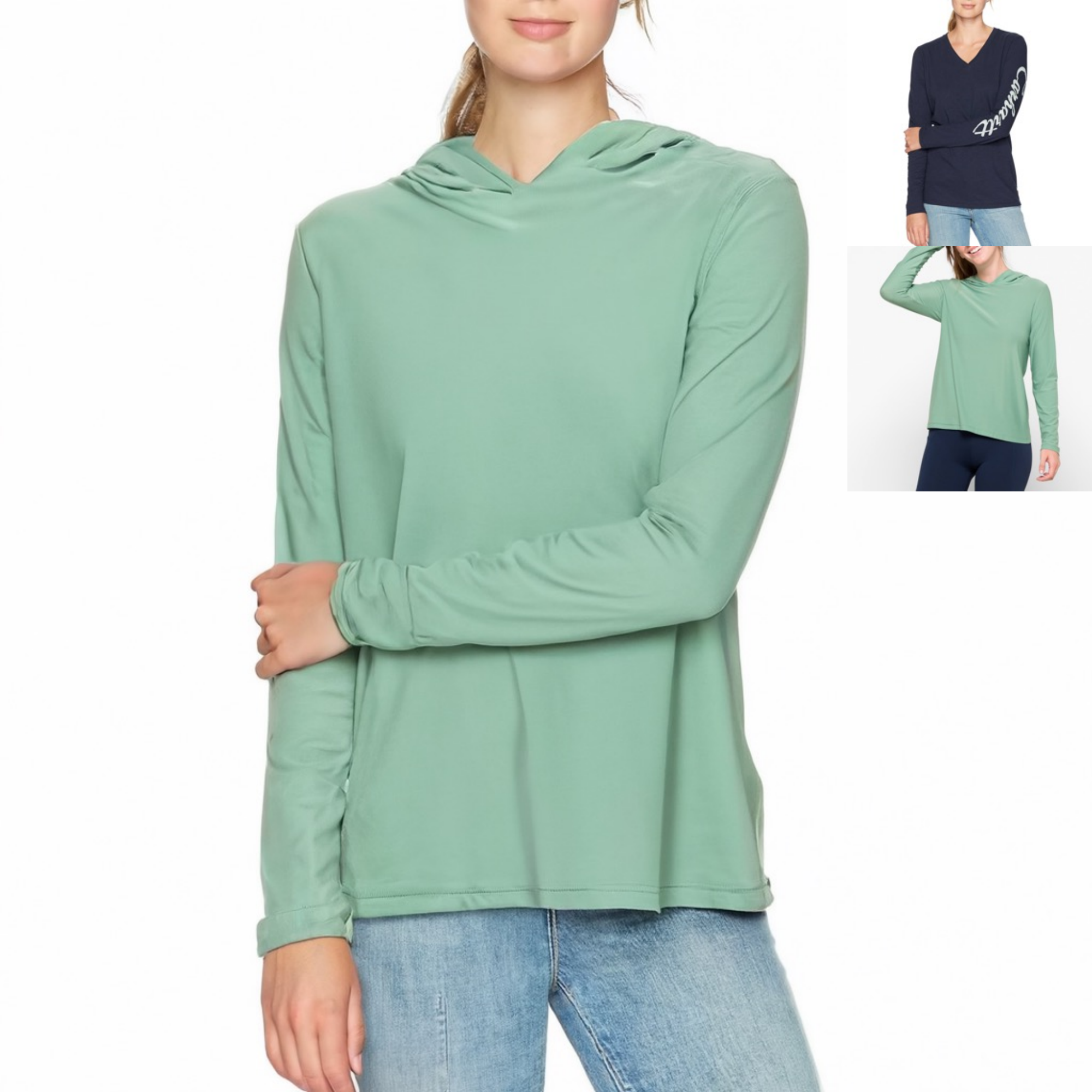}
\end{center}
  \caption{Larger version of teaser.}
\label{fig:suppl_teaser_woman2}
\end{figure*}

\begin{figure*}
\begin{center}
  \includegraphics[width=1.0\linewidth]{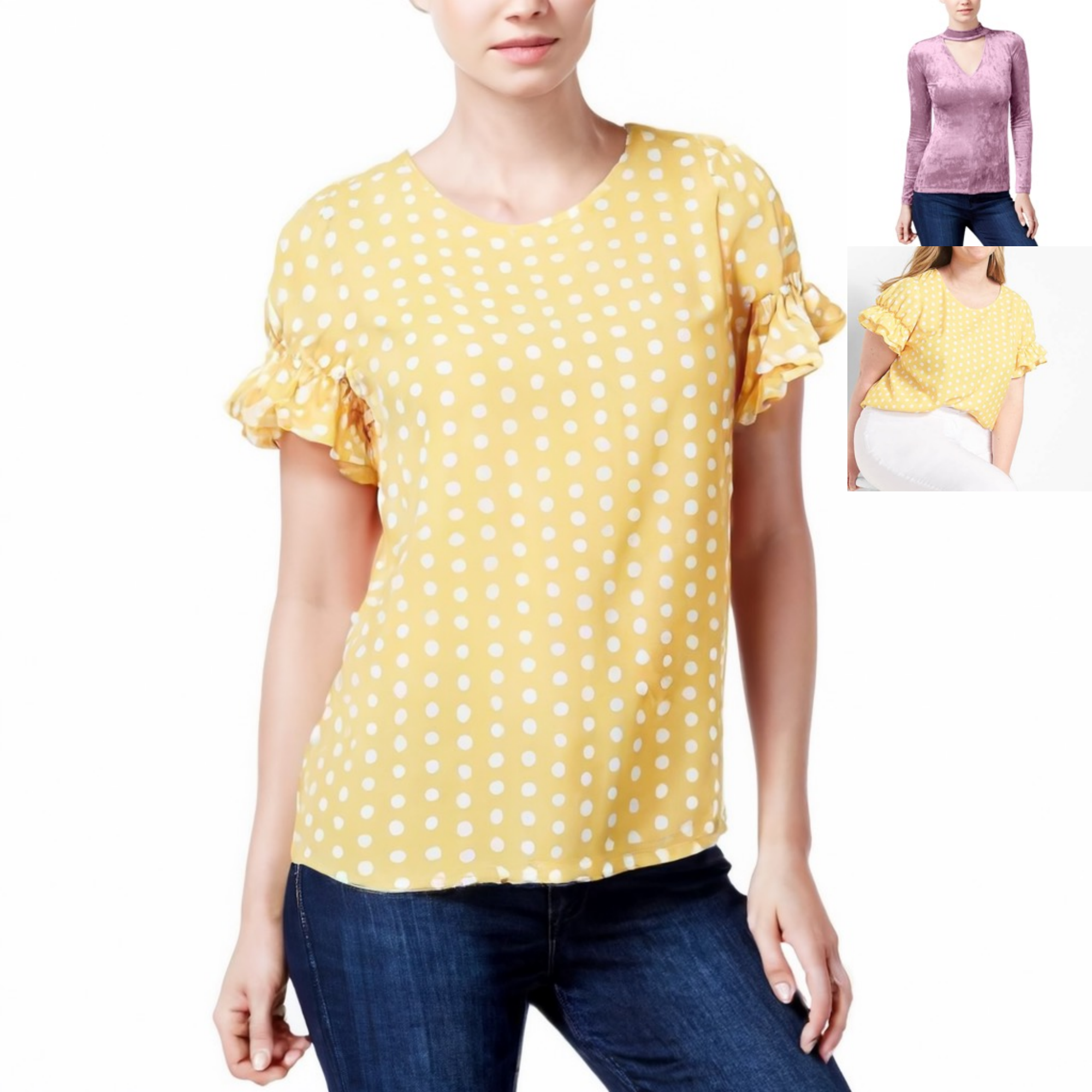}
\end{center}
  \caption{Larger version of teaser.}
\label{fig:suppl_teaser_woman3}
\end{figure*}

\begin{figure*}
\begin{center}
  \includegraphics[width=1.0\linewidth]{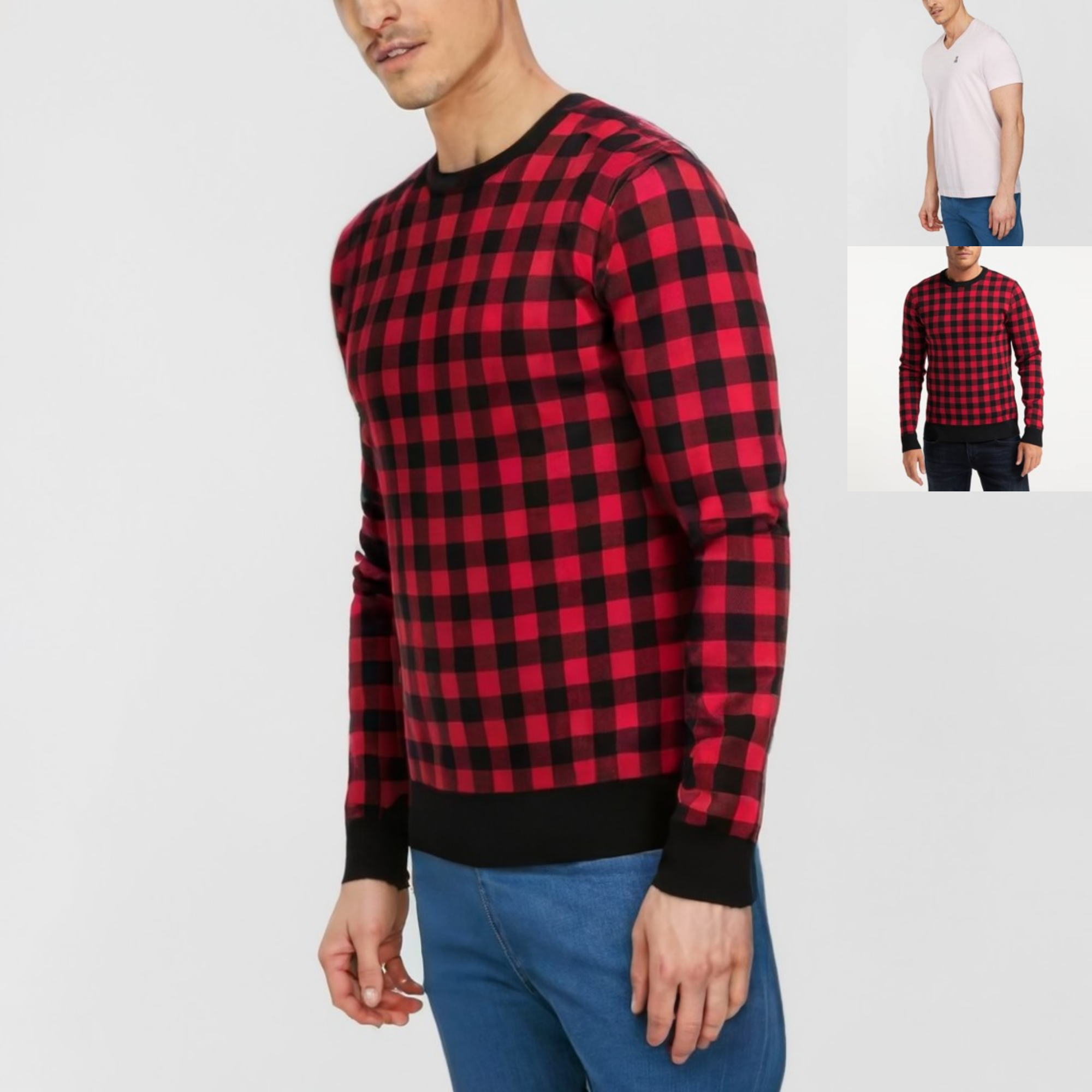}
\end{center}
  \caption{Larger version of teaser.}
\label{fig:suppl_teaser_man1}
\end{figure*}

\begin{figure*}
\begin{center}
  \includegraphics[width=1.0\linewidth]{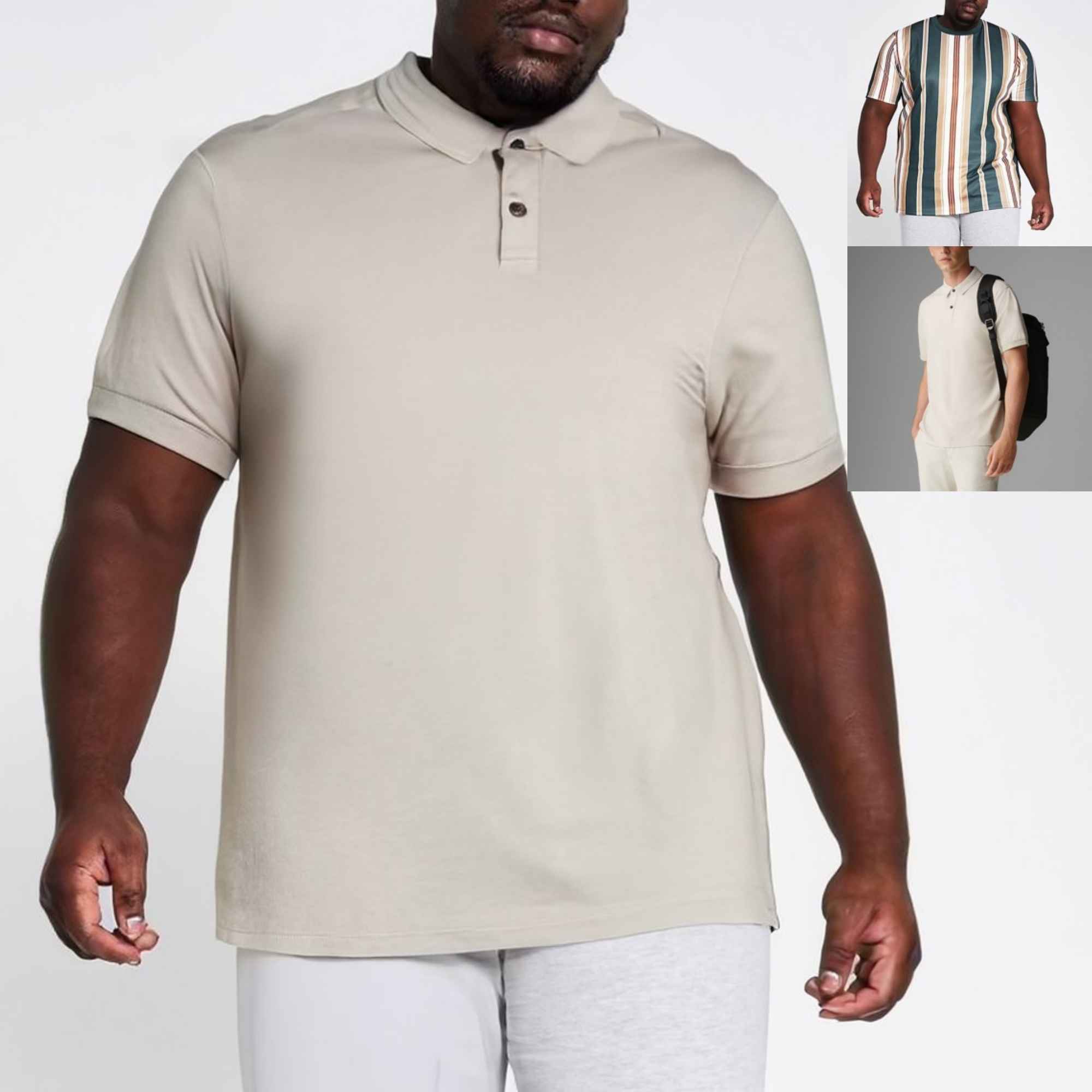}
\end{center}
  \caption{Larger version of teaser.}
\label{fig:suppl_teaser_man2}
\end{figure*}

\begin{figure*}
\begin{center}
  \includegraphics[width=1.0\linewidth]{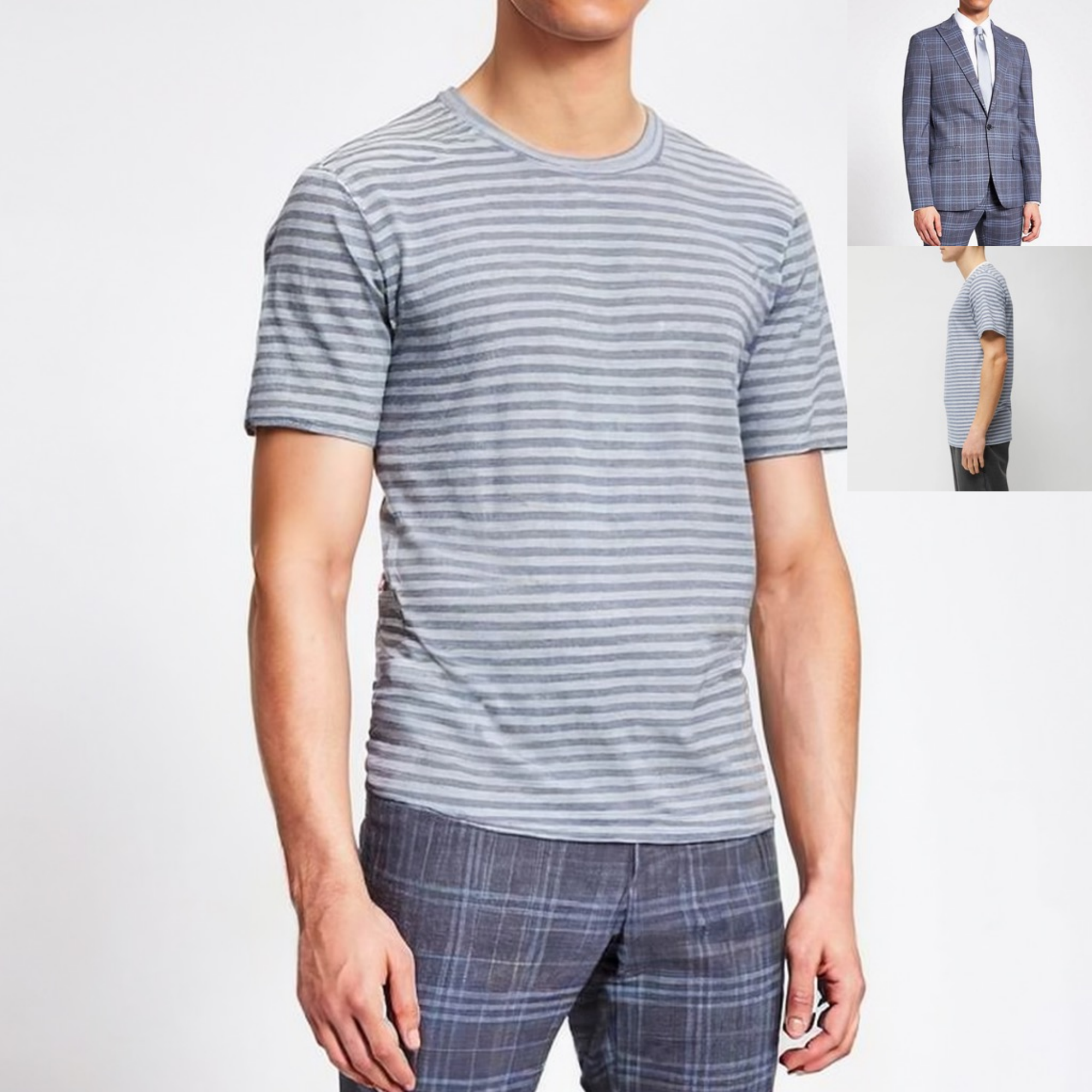}
\end{center}
  \caption{Larger version of teaser.}
\label{fig:suppl_teaser_man3}
\end{figure*}

Finally, Fig.~\ref{fig:suppl_teaser_woman1} to \ref{fig:suppl_teaser_man3} provide zoom-in visualization for Fig.~\ref{fig:teaser} of the main paper, demonstrating high quality results of our method.

\begin{table}
\setlength\tabcolsep{1.5pt}
\centering
\begin{tabular}{ |c|c|c|c|}
\hline
 & SDAFN~\cite{bai2022single} & Ours & Hard to tell \\ 
\hline 
Random  &  5.24\% &  \textbf{77.83\%}  &  16.93\% \\
Challenging &  3.96\% & \textbf{93.99\%} & 2.05\% \\
\hline
\end{tabular}
\caption{User study comparing SDAFN~\cite{bai2022single} to our method at $256 \mathord\times\mathord256$ resolution.}
\label{table:user_study_256}
\end{table}

\begin{table}
\centering
\begin{tabular}{ |c|cc|cc|  }
\hline
\multicolumn{1}{|c}{Test datasets} & \multicolumn{2}{|c}{Ours } & \multicolumn{2}{|c|}{VITON-HD}\\
\hline
Methods & FID $\downarrow$ & KID $\downarrow$ & FID $\downarrow$ & KID $\downarrow$\\ 
\hline 
Ablation 1 & 15.691   &  7.956 & 25.093 & 12.360 \\
Ablation 2 &  14.936  & 7.235 & 28.330 & 17.339  \\
Ours & \textbf{13.447} & \textbf{6.964} & \textbf{23.352} & \textbf{10.838} \\
 \hline
\end{tabular}
\caption{Quantitative comparison for ablation studies. We compute FID and KID on our 6K test set and VITON-HD's unpaired test set. The KID is scaled by 1000 following \cite{karras2020training}.}
\label{table:quantitative_ablation}
\end{table}

\begin{table}
\centering
\begin{tabular}{ |c|cc|cc|  }
\hline
\multicolumn{1}{|c}{Test datasets} & \multicolumn{2}{|c}{Ours } & \multicolumn{2}{|c|}{VITON-HD}\\
\hline
Train set size & FID $\downarrow$ & KID $\downarrow$ & FID $\downarrow$ & KID $\downarrow$\\ 
\hline 
10K & 16.287   &  8.975 & 25.040 & 11.419 \\
100K &  14.667  & 7.073 & 23.983 & \textbf{10.732}  \\
4M & \textbf{13.447} & \textbf{6.964} & \textbf{23.352} & 10.838 \\
 \hline
\end{tabular}
\caption{Quantitative results for the effects of the training set size. We compute FID and KID on our 6K test set and VITON-HD's unpaired test set. The KID is scaled by 1000 following \cite{karras2020training}.}
\label{table:quantitative_dataset_size}
\end{table}

\end{document}